\documentclass[conference]{IEEEtran}
\usepackage{times}

\usepackage[numbers]{natbib}
\usepackage{multicol}
\usepackage[bookmarks=true]{hyperref}
\usepackage{xspace}
\providecommand{\eg}{\textit{e.g.}}
\providecommand{\ie}{\textit{i.e.}}
\usepackage{amsmath}
\usepackage{tabularx}
\usepackage{amsfonts} 
\usepackage{algorithm}          
\usepackage{algpseudocode}      
\usepackage{amsmath}            
\usepackage{amssymb}            
\usepackage{adjustbox}
\usepackage{threeparttable} 
\usepackage{booktabs} 
\usepackage{multirow} 
\usepackage{graphicx} 
\usepackage{makecell}
\usepackage{array}
\usepackage{diagbox}
\usepackage{marvosym}
\usepackage{caption}
\usepackage[table]{xcolor}
\definecolor{bgpink}{RGB}{255, 235, 235} 
\definecolor{bgblue}{RGB}{235, 235, 255} 
\definecolor{graycol}{gray}{0.95}        
\definecolor{textgreen}{RGB}{46, 139, 87} 
\definecolor{textred}{RGB}{220, 20, 60}   

\let\titleold\title
\renewcommand{\title}[1]{\titleold{#1}\newcommand{\thetitle}{#1}}
\def\maketitlesupplementary
{
    \newpage
    \begin{center}
        \Large
        \textbf{Towards Long-Lived Robots: \\ Continual Learning VLA Models \\ via Reinforcement Fine-Tuning}\\
        \vspace{0.5em}Supplementary Material \\
        \vspace{1.0em}
    \end{center}
}

\begin{document}

\title{Towards Long-Lived Robots: Continual Learning VLA Models via Reinforcement Fine-Tuning}


\author{\authorblockN{Yuan Liu\textsuperscript{1,2}\textsuperscript{\dag}, 
Haoran Li\textsuperscript{1,3,4~\Letter}, 
Shuai Tian\textsuperscript{1,3}, 
Yuxing Qin\textsuperscript{1,3}, 
Yuhui Chen\textsuperscript{1,3}, 
Yupeng Zheng\textsuperscript{1,3}, \\
Yongzhen Huang\textsuperscript{2}, 
Dongbin Zhao\textsuperscript{1,3,4}}
\authorblockA{\textsuperscript{1}SKL-MAIS, Institute of Automation, Chinese Academy of Sciences (CASIA), Beijing, China}
\authorblockA{\textsuperscript{2}School of Artificial Intelligence, Beijing Normal University, Beijing, China}
\authorblockA{\textsuperscript{3}School of Artificial Intelligence, University of Chinese Academy of Sciences, Beijing, China}
\authorblockA{\textsuperscript{4}Beijing Academy of Artificial Intelligence, Beijing, China}
}


%


\twocolumn[{%
\renewcommand\twocolumn[1][]{#1}%
\maketitle
    \begin{center}
            \vspace{-0.7cm}
            \captionsetup{type=figure}
            \includegraphics[width=\textwidth]{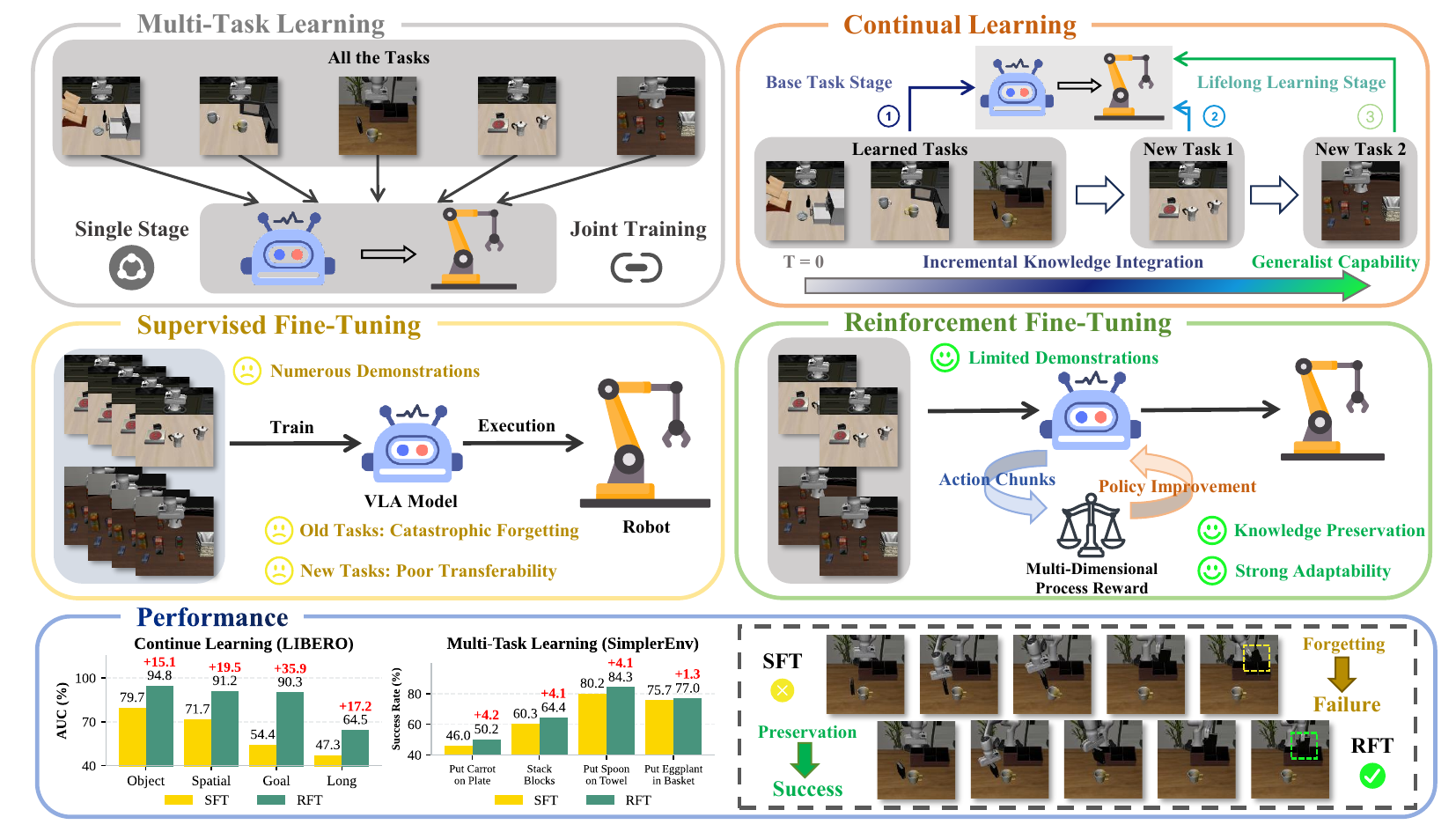}
            \vspace{-5mm}
            \caption{\textbf{Overview of VLA post-training.} This phase involves single-stage multi-task adaptation and incremental continual learning. Addressing the substantial data dependence and susceptibility to catastrophic forgetting inherent in SFT, we introduce LifeLong-RFT, which combines on-policy RL with the multi-dimensional process reward mechanism.}
            \label{fig:teaser}
    \end{center}
}]

\begingroup 
\renewcommand\thefootnote{}
\footnotetext[1]{\textsuperscript{\dag}~Work done during an internship at CASIA. \textsuperscript{\Letter}~Corresponding author.} 
\endgroup 

\begin{abstract}
Pretrained on large-scale and diverse datasets, VLA models demonstrate strong generalization and adaptability as general-purpose robotic policies.
However, Supervised Fine-Tuning (SFT), which serves as the primary mechanism for adapting VLAs to downstream domains, requires substantial amounts of task-specific data and is prone to catastrophic forgetting.
%
%
%
To address these limitations, we propose LifeLong-RFT, a simple yet effective Reinforcement Fine-Tuning (RFT) strategy for VLA models independent of online environmental feedback and pre-trained reward models.
By integrating chunking-level on-policy reinforcement learning with the proposed multi-dimensional process reward mechanism, LifeLong-RFT quantifies the heterogeneous contributions of intermediate action chunks across three dimensions to facilitate policy optimization.
%
%
Specifically, (1) the Quantized Action Consistency Reward (QACR) ensures accurate action prediction within the discrete action space;
(2) the Continuous Trajectory Alignment Reward (CTAR) aligns decoded continuous action chunks with reference trajectories to ensure precise control;
(3) the Format Compliance Reward (FCR) guarantees the structural validity of outputs.
%
%
Comprehensive experiments across SimplerEnv, LIBERO, and real-world tasks demonstrate that LifeLong-RFT exhibits strong performance in multi-task learning.
Furthermore, for continual learning on the LIBERO benchmark, our method achieves a 22\% gain in average success rate over SFT, while effectively adapting to new tasks using only 20\% of the training data.
Overall, our method provides a promising post-training paradigm for VLAs.
The project page is available at \href{https://yuan-liu-lifelong-rft.github.io}{https://yuan-liu-lifelong-rft.github.io}.
%
%
%
%

\end{abstract}

\IEEEpeerreviewmaketitle

\section{Introduction}

Vision-Language-Action (VLA) models trained on large-scale datasets are progressively emerging as a pivotal approach for achieving generalist robot policies~\cite{black2024pi_0,intelligence2025pi_,hung2025nora, chen2023robogpt}. Despite these advances, adapting VLA models to new tasks via supervised fine-tuning (SFT) remains challenging, as illustrated in Fig.~\ref{fig:teaser}. First, SFT typically requires a substantial amount of task-specific data, limiting the ability of VLA models to rapidly adapt in low-data or few-shot settings. Second, SFT often leads to catastrophic forgetting, where learning new skills degrades previously acquired knowledge. These issues prevent the SFT from supporting the evolution of VLA into long-lived agents capable of continually acquiring new skills.
%

These two challenges are not independent~\cite{tao2020few}: improving data-efficient adaptation often exacerbates forgetting, while preserving prior knowledge restricts effective learning from limited new data. Achieving an effective balance between plasticity and stability is essential for robots to learn from limited data without erasing prior knowledge. In earlier work based on specialized models, this trade-off was widely viewed as intrinsic~\cite{ma2025latest}, motivating solutions based on task-specific adapters or handcrafted features~\cite{lee2024incremental,liu2023tail,wang2024sparse,meng2025preserving}. With the emergence of foundation models, representations learned from massive and diverse datasets exhibit substantially improved transferability, reshaping the plasticity–stability dilemma yet not eliminating it~\cite{park2024pre}. While such representations significantly reduce the data requirements for learning new tasks, directly applying SFT still results in severe catastrophic forgetting~\cite{luo2025empirical}. Consequently, continual learning techniques developed for specialized models are often reused to mitigate forgetting in foundation models~\cite{wan2024lotus,wu2025continually,song2025few}. However, these techniques struggle to scale to VLA settings involving both massive tasks and a high-capacity parameterization.


In contrast to SFT, which learns from the annotated datasets, recent advances in large language models suggest that on-policy reinforcement learning (RL), which updates the model using samples drawn from its current distribution, can exhibit stronger robustness to forgetting~\cite{shenfeld2025rl,chen2025retaining,lai2025reinforcement}. This observation raises an important question for robotics: \emph{can on-policy reinforcement learning be leveraged to enable continual adaptation of VLA foundation models, supporting their evolution into long-lived agents?} A central challenge in answering this question lies in designing efficient, reliable, and scalable reward signals for reinforcement fine-tuning of VLA models.


Existing approaches to reinforcement fine-tuning VLA models rely primarily on two categories of reward signals. The first uses environment-provided ground-truth rewards~\cite{liu2025can,li2025simplevla,tan2025interactive,lyu2025reinforcement,chen2025pirl}, which are typically available only in simulation and depend on privileged information. Such methods face significant barriers in real-world deployment due to the sim-to-real gap and the difficulty of computing rewards without access to privileged state. The second category employs model-based reward estimation, such as predicting task success ~\cite{sontakke2023roboclip,luo2025precise,chen2025conrft,zhang2025reinforcing}, task progress~\cite{lu2025vla,chen2025tgrpo,zhai2025vision,lee2026roborewardgeneralpurposevisionlanguagereward}, or distance-based dense rewards~\cite{escontrela2023video,huang2024diffusion,chen2025tevir,hung2025nora,fei2025srpo}. However, inaccuracies and generalization errors in reward models make these approaches highly vulnerable to reward hacking~\cite{christiano2017deep}. Moreover, both categories require extensive interaction with an environment—whether simulators~\cite{li2025simplevla,liu2025can}, world models~\cite{wmpo2025zhu,hung2025nora}, or real robots~\cite{chen2025conrft,pan2026sop}—resulting in prohibitively high training costs when scaling to large task sets. Importantly, existing methods predominantly optimize performance on the fine-tuning tasks themselves, while largely neglecting the continual learning properties required for long-lived VLA agents.

In this work, we propose a simple yet effective post-training paradigm for VLA models named \textbf{LifeLong-RFT}. By designing a multi-dimensional process reward mechanism, we enable chunking-level on-policy reinforcement fine-tuning without requiring interaction with the environment.
Specifically, we decompose this mechanism into three dimensions to provide comprehensive rewards.
First, we introduce the \textbf{Q}uantized \textbf{A}ction \textbf{C}onsistency \textbf{R}eward (QACR).
Given that VLAs are built upon VLM backbones to generate discrete action tokens, QACR ensures precise prediction within the quantized action space by measuring the consistency between predicted and target tokens.
%
%
%
%
%
Second, we design the \textbf{C}ontinuous \textbf{T}rajectory \textbf{A}lignment \textbf{R}eward (CTAR).
While QACR ensures accuracy within the quantized action space, physical execution necessitates alignment with continuous trajectories.
To this end, CTAR utilizes decoded action chunks to compute chunking-level rewards based on spatial deviations from reference trajectories, incentivizing the model to explore optimal motions.
Third, we introduce the \textbf{F}ormat \textbf{C}ompliance \textbf{R}eward (FCR).
Due to the generative diversity of VLA backbones, the model is prone to producing structurally invalid outputs (\eg, mismatched action dimensions and inconsistent prediction horizons).
To mitigate this instability, the FCR functions as a binary reward that promotes adherence to valid formats, ensuring action executability and enhancing inference efficiency.
%
%

%
Our main contributions are summarized as follows:
1) We propose \textbf{LifeLong-RFT}, a post-training strategy that integrates chunking-level on-policy reinforcement learning with the multi-dimensional process reward.
This approach enables VLAs to continually master new tasks with limited demonstrations while preserving original capabilities.

%
%
%
%
2) The multi-dimensional process reward mechanism comprises the \textbf{Q}uantized \textbf{A}ction \textbf{C}onsistency \textbf{R}eward (QACR), the \textbf{C}ontinuous \textbf{T}rajectory \textbf{A}lignment \textbf{R}eward (CTAR), and the \textbf{F}ormat \textbf{C}ompliance \textbf{R}eward (FCR), quantifying the heterogeneous contributions of intermediate action chunks across three dimensions to facilitate policy optimization.

%
%
%
3) Comprehensive experiments across both simulated and real-world tasks demonstrate LifeLong-RFT's superior performance in multi-task learning. Notably, for continual learning on LIBERO, our method achieves a 22\% improvement in average success rate over SFT, facilitating efficient adaptation to novel tasks with only 20\% of the training data.
%

%

\section{Related Work}

\subsection{Vision-Language-Action Models}
Representing a paradigm shift, VLA models diverge from the traditional hierarchical architecture in favor of an end-to-end learning approach, directly mapping multimodal perceptual inputs to robot control actions~\cite{brohan2022rt,zitkovich2023rt}. Generally, these models can be categorized into two streams based on their action representations: Discrete Action Models~\cite{pertsch2025fast,lee2025molmoact,qu2025spatialvla,kim2024openvla, RoboBrain1.0} and Continuous Action Models~\cite{kim2025fine,intelligence2025pi_,team2024octo,huang2025thinkact,zheng2025x}. Discrete Action Models typically utilize VLM backbones~\cite{bai2025qwen2, beyer2024paligemma} to generate discrete action tokens autoregressively for executing manipulation tasks. In contrast, Continuous Action Models explore the integration of diffusion policies~\cite{team2024octo, chen2023boosting} or flow matching~\cite{black2024pi_0} with VLMs to directly output continuous actions, achieving dexterous control. Building on these architectures, current models typically leverage large-scale pretraining to acquire manipulation priors, followed by SFT to adapt to specific downstream tasks. Nevertheless, despite demonstrating promising performance, this SFT-based post-training paradigm remains limited by the need for substantial amounts of task-specific data and is prone to catastrophic forgetting.
%
%
%
%
%
%

\subsection{Reinforcement Fine-tuning for VLA Models}
To further enhance the robustness and self-refinement capabilities of VLAs, recent research increasingly explores reinforcement fine-tuning strategies. Current strategies primarily comprise three paradigms: simulation-based, real-world-based, and world model-driven approaches. Simulation-based methods~\cite{lyu2025reinforcement,zang2025rlinf,li2025simplevla,lu2025vla,chen2025tgrpo} benefit from large-scale parallelization, significantly enhancing sample efficiency, and leverage privileged states to construct dense rewards. However, constrained by the sim-to-real gap, these approaches face challenges in deployment within the physical world. Real-world-based strategies~\cite{chen2025conrft,guo2025improving,yuan2024policy,zhang2025rewind} effectively enhance model generalization through online adaptation to physical environments. Nevertheless, such methods often involve prohibitive human costs and struggle with the acquisition of rewards. Notably, frontier research~\cite{hung2025nora,fei2025srpo,li2025vla,xiao2025world,jiang2025world4rl} employs world models for VLA reinforcement fine-tuning. By leveraging the capability of future state prediction, this approach provides dense reward signals for policy optimization. However, the inherent prediction errors of world models increase the susceptibility of VLAs to reward hacking. Overall, these methods necessitate extensive environmental interaction, limiting their scalability due to high training costs.
%

%
%
%
%
%
%
%
%
\subsection{Continual Learning in Robotics}
Continual learning in robotics aims to construct generalist policies capable of adapting to dynamic environmental changes~\cite{beck2023survey,beck2025tutorial} while retaining existing capabilities.
Several studies~\cite{mallya2018packnet,liu2023tail,lee2024incremental,wang2024sparse,li2019learn,rao2019continual,xu2025speci} address forgetting by allocating specific parameters to each new learning phase. Additionally, alternative approaches depend on task decomposition via clustering or multistage learning~\cite{wan2024lotus,zhu2022bottom, meng2025preserving}.
%
With the advent of VLAs, recent research focuses on enabling their continual adaptation.
Along this line, MergeVLA~\cite{fu2025mergevla} introduces a model-merging paradigm, aiming to achieve efficient skill expansion by resolving parameter conflicts during the fusion of multi-expert models.
On the other hand, Stellar VLA~\cite{wu2025continually} constructs a knowledge-driven continual imitation learning framework, effectively mitigating catastrophic forgetting.
In contrast to the aforementioned methods, we integrate on-policy reinforcement learning with the proposed multi-dimensional process reward mechanism to effectively adapt to new tasks while retaining previously learned knowledge.
%
%

\section{Problem Formulation and Preliminaries}
%
\textbf{VLA and Post-Training.} The goal of VLA modeling is to learn a general-purpose robotic policy $\pi_{\theta}(\mathbf{a} | o, l),$
which maps observations \(o\) and natural language instructions \(l\) to robot actions \(\mathbf{a}\).
In practice, a VLA model is first \emph{pretrained} on large-scale and diverse datasets to acquire rich semantic understanding and transferable representations.
The pretrained parameters \(\theta\) are then \emph{post-trained} using task-specific data to adapt the action outputs \(\mathbf{a}\) to the target robot embodiment and downstream tasks.

\textbf{Continual Learning.} SFT remains the primary approach for post-training VLA models. However, SFT primarily optimizes performance on the tasks present in the current training dataset, while largely overlooking the degradation of previously acquired capabilities. In real-world settings, a long-lived agent is expected to acquire new skills while retaining the skills learned earlier, a requirement commonly referred to as continual learning. Formally, this involves an agent learning from a sequence of tasks  $\{\mathcal{T}_k\}_{k=1}^{\infty}$, where each task $\mathcal{T}_k$ is associated with $N$ expert demonstrations $\{\tau_k^n\}_{n=1}^{N}$. Unlike a single adaptation stage, which assumes concurrent access to all task data, CL necessitates continuous knowledge acquisition under the constraint of limited access to historical data.
%
%
%
%
%
%
%
%
\begin{figure*}[tp]
    \centering
    \includegraphics[width=\linewidth]{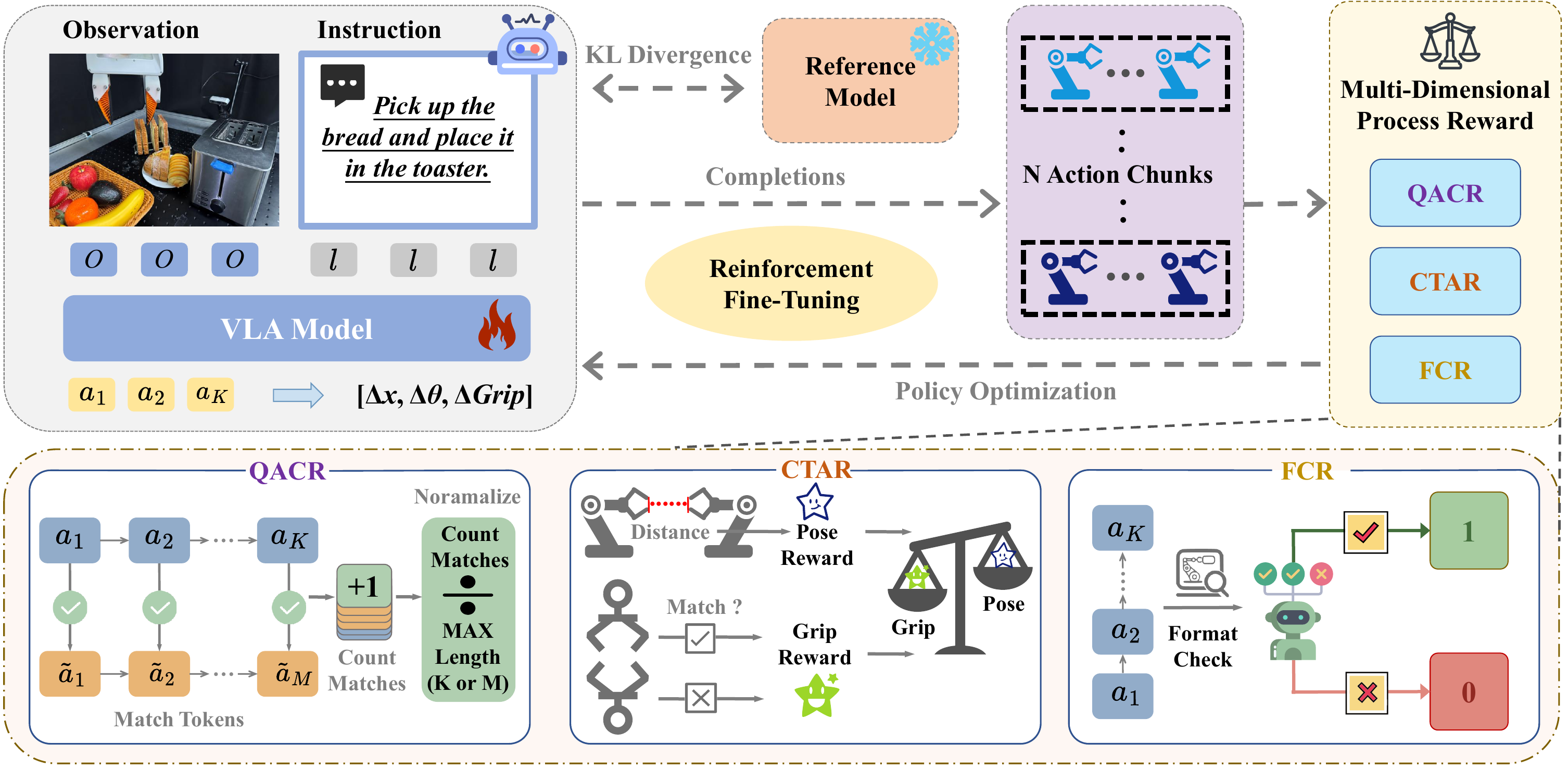}
    \captionof{figure}{
    \textbf{Overview of the proposed LifeLong-RFT.} This strategy integrates the chunking-level on-policy reinforcement learning algorithm with the multi-dimensional process reward mechanism to facilitate policy optimization.
    }
\label{fig:method}
\vspace{-0.3cm}
\end{figure*}
%

\textbf{On-Policy Reinforcement Learning.} 
While SFT can efficiently improve performance on the currently targeted tasks, it often leads to rapid degradation of previously acquired capabilities, a phenomenon commonly known as catastrophic forgetting.
In contrast, recent findings~\cite{shenfeld2025rl,chen2025retaining,lai2025reinforcement} in LLMs suggest that on-policy reinforcement learning exhibits stronger resistance to forgetting. Unlike SFT, which relies on fixed annotated datasets, on-policy reinforcement learning updates the policy using self-generated answers and optimizes the expected return over these answers.


\section{Method}
%
To support the evolution of VLAs into long-lived agents capable of continually acquiring new skills, we propose LifeLong-RFT, a reinforcement fine-tuning strategy illustrated in Fig.~\ref{fig:method}.
This strategy integrates chunking-level on-policy reinforcement learning with the proposed multi-dimensional process reward mechanism, which quantifies the heterogeneous contributions of intermediate action chunks across three dimensions without requiring environment interaction.
%
%

\subsection{Chunking-Level On-Policy Reinforcement Learning}
Most existing on-policy RL approaches~\cite{liu2025can,li2025simplevla,tan2025interactive,chen2025pirl} for VLA post-training optimize model parameters by collecting full trajectories and relying on environment-provided rewards. While such methods can achieve strong performance, they require extensive interaction with the environment during training, leading to high training costs and limiting scalability to large-scale and multi-task settings. To eliminate the need for environment interaction, we adopt a simple alternative: instead of evaluating actions along complete trajectories, we evaluate each action chunk sampled by the VLA model independently, thereby removing the dependency on environment interaction. In this work, we employ Group Relative Policy Optimization (GRPO)~\cite{shao2024deepseekmath}. In contrast to conventional algorithms such as PPO~\cite{schulman2017proximal}, which rely on an explicit critic network, GRPO estimates advantages via group-wise comparisons of sampled outputs, thereby considerably reducing computational overhead.
Specifically, for each observation $o$ and instruction $l$, a group of $G$ action outputs $\{\mathbf{a}_i\}_{i=1}^{G}$ is first sampled from the old policy $\pi_{\theta_{\text{old}}}(\mathbf{a} | o, l)$.
Then, corresponding rewards \(\{r_i\}_{i=1}^{G}\) are computed via task-specific reward functions.
Based on the mean and standard deviation of intra-group rewards, the relative advantage $A_i$ for each output is computed as follows:
\begin{equation}
A_i = \frac{r_i - \mathrm{mean}(\{r_1, \dots, r_G\})}{\mathrm{std}(\{r_1, \dots, r_G\})}.
\end{equation}
Given the advantage $A_i$, the policy parameters $\theta$ are optimized by maximizing the following objective:
\begin{equation}
\begin{aligned}
    J_{\text{GRPO}}(\theta) &= \mathbb{E}_{ (o, l) \sim \mathcal{B}, \{ \mathbf{a}_i \}_{i=1}^{G} \sim \pi_{\theta_{\text{old}}}(\cdot | o, l) } \\
    & \quad \frac{1}{G} \sum_{i=1}^{G} \{ \min [ \frac{\pi_{\theta}(\mathbf{a}_{i} | o, l)}{\pi_{\theta_{\text{old}}}(\mathbf{a}_{i} | o, l)} A_{i}, \\
    & \quad \text{clip} \left( \frac{\pi_{\theta}(\mathbf{a}_{i} | o, l)}{\pi_{\theta_{\text{old}}}(\mathbf{a}_{i} | o, l) }, 1 - \epsilon, 1 + \epsilon \right) A_{i} ] \\
    & \quad - \gamma D_{KL} \left[ \pi_{\theta} || \pi_{\text{ref}} \right] \},
\end{aligned}
\label{GRPO}
\end{equation}
%
where $\mathcal{B}$ denotes the dataset of expert demonstrations, each comprising an observation $o$ and a language instruction $l$.
%
To stabilize the training process, $\text{clip}$ constrains the policy probability ratio, $\frac{\pi_{\theta}(\mathbf{a}_{i} | o, l)}{\pi_{\theta_{\text{old}}}(\mathbf{a}_{i} | o, l)}$, within $[1 - \epsilon, 1 + \epsilon]$.
Furthermore, $\gamma$ modulates the strength of the KL divergence regularization term $D_{KL} \left[ \pi_{\theta} || \pi_{\text{ref}} \right]$, effectively preventing the new policy $\pi_{\theta} $ from deviating excessively from the reference policy $\pi_{\text{ref}}$. Building upon this formulation, the construction of an efficient and verifiable reward $r_i$ becomes the key to optimization.

\subsection{Multi-Dimensional Process Reward}
To effectively guide the on-policy reinforcement learning process without requiring environment interaction, we design the multi-dimensional process reward mechanism. This mechanism decomposes the assessment of action chunks into three complementary dimensions, bridging discrete token generation and continuous robotic control. In this section, we detail the designs of the three dimension-specific rewards.
\subsubsection{Quantized Action Consistency Reward}
Built upon VLM backbones, contemporary VLAs~\cite{hung2025nora_1,kim2024openvla,pertsch2025fast} interpret language instructions and multi-modal observations to generate action tokens.
%
%
This paradigm necessitates designing a specialized reward function to assess the consistency between generated tokens and the ground truth, facilitating accurate prediction within the quantized action space.
For this purpose, we propose the Quantized Action Consistency Reward (QACR) function, as shown in Algorithm ~\ref{alg:qacr_function}.
%

%
%
%
First, we perform a format check on the model generations to verify their compliance with the predefined specifications of the action tokenizer Fast+~\cite{pertsch2025fast} (\ie, the action chunk size and action dimension).
%
Only validated generations proceed to the subsequent consistency assessment stage, while those failing the verification receive a reward of zero.
%
Second, we compute the consistency reward by position-wise matching between the predicted action token sequence $\mathbf{a} = \{a_u\}_{u=1}^U$ and its ground-truth counterpart $\tilde{\mathbf{a}} = \{\tilde{a}_v\}_{v=1}^V$, which is formally defined as:
%
\begin{equation}
\label{eq:qacr_refined}
\text{QACR} = \begin{cases} 
    \displaystyle \frac{\sum_{\ell=1}^{\min(U, V)} \mathbb{I}(a_\ell = \tilde{a}_\ell)}{\max(U, V)}, & \text{if valid} \\[1.5em]
    0, & \text{otherwise}
\end{cases}
\end{equation}
where $\mathbb{I}(\cdot)$ is the indicator function that returns 1 when the predicted action token $a_\ell$ matches the ground-truth $\tilde{a}_\ell$, and 0 otherwise.
Additionally, the term ``valid'' indicates that the predicted sequence satisfies the decoding requirements of the Fast+ tokenizer.
%
%
Based on this formulation, QACR provides a robust assessment for sequence consistency.
%
%
%
\begin{algorithm}[tp]
\caption{Pseudo code of the QACR function} 
\label{alg:qacr_function}
\textbf{Input:} Predicted action token sequence $\mathbf{a} = \{a_u\}_{u=1}^U$; Ground-truth action token sequence $\tilde{\mathbf{a}} = \{\tilde{a}_v\}_{v=1}^V$ \\[-1.1em]
\begin{algorithmic}[1]
    \State $\text{is\_valid} \gets \Call{FormatCheck}{\mathbf{a}}$
    
    \If{$\text{is\_valid} = \text{False}$}
        \State $\text{QACR} \gets 0$ \Comment{Invalid format yields zero reward}
    \Else        
        \State $\text{QACR} \gets \dfrac{\sum_{\ell=1}^{\min(U, V)} \mathbb{I}(a_\ell = \tilde{a}_\ell)}{\max(U, V)}$
    \EndIf
    \State \textbf{return} \text{QACR}
\end{algorithmic}
\textbf{Output:} The QACR score $\in [0, 1]$
\end{algorithm}
%
%
\subsubsection{Continuous Trajectory Alignment Reward}
%
%
While QACR ensures accuracy within the quantized action space, physical execution necessitates alignment with continuous trajectories.
%
%
To address this, we introduce the Continuous Trajectory Alignment Reward (CTAR).
This mechanism assesses the spatial alignment between decoded continuous action chunks and reference trajectories, providing dense feedback to facilitate dexterous manipulation.
%
%
The implementation of this reward function is outlined in Algorithm ~\ref{alg:ctar_function}.

Consistent with QACR, we first conduct format verification on the predicted action token sequence $\mathbf{a} = \{a_u\}_{u=1}^U$.
Only sequences that pass this verification proceed to the subsequent reward calculation, while invalid ones are directly assigned a zero reward.
Subsequently, we utilize the Fast+~\cite{pertsch2025fast} tokenizer to decode the predicted action tokens into the continuous action chunk $\mathbf{y}$, comprising a sequence of $H$ actions.
%
%
For the action chunk $\mathbf{y}$, the action vector $\mathbf{y}_t$ at each time step consists of a pose component $\mathbf{y}_t^{\text{pose}}$ and a gripper component $\mathbf{y}_t^{\text{grip}}$. 
Here, $\mathbf{y}_t^{\text{pose}}$ represents the end-effector pose (or joint angles) of the robot at step $t$, while $\mathbf{y}_t^{\text{grip}}$ indicates the gripper's open-close state.
Based on this, we decompose the calculation of CTAR into the following steps:
\textbf{(1)} To encourage precise pose alignment, we formulate the pose reward $r_t^{\text{pose}}$ as an exponentially decaying function of the error relative to the ground truth.
Specifically, we compute the normalized L1 distance $d_t$ between the predicted pose vector $\mathbf{y}_t^{\text{pose}}$ and the ground truth $\tilde{\mathbf{y}}_t^{\text{pose}}$.
Based on this error, we apply an exponential decay function $\exp(-\alpha \cdot d_t)$ to convert it into a reward signal, where the hyperparameter $\alpha$ regulates the sensitivity to pose deviation.
%
%
\textbf{(2)} To incentivize precise gripper actuation, we employ a binary reward $r_t^{\text{grip}}$. 
This reward is defined as an indicator function $\mathbb{I}(\cdot)$ that assigns a value of 1 when the predicted gripper state $\mathbf{y}_t^{\text{grip}}$ matches the ground truth $\tilde{\mathbf{y}}_t^{\text{grip}}$, and 0 otherwise.
%
%
\textbf{(3)} Finally, the normalized CTAR is computed by averaging the weighted combination of pose and grip rewards over the action chunk size $H$, formally defined as:
%
%
%
%
\begin{equation}
    \text{CTAR} = \begin{cases} 
        \displaystyle \frac{1}{H} \sum_{t=1}^{H} \left( \beta \cdot r_t^{\text{pose}} + (1 - \beta) \cdot r_t^{\text{grip}} \right), & \text{if valid} \\[1.2em]
        0, & \text{otherwise}
    \end{cases}
\end{equation}
where the hyperparameter $\beta \in [0, 1]$ modulates the relative importance of the pose reward $r_t^{\text{pose}}$ and the gripper reward $r_t^{\text{grip}}$ at each time step $t$.
%
%
%
In conclusion, the CTAR function establishes a dense reward mechanism by quantifying the prediction discrepancies in both robot poses and gripper states.
%
%

%
\begin{algorithm}[tp]
\caption{Pseudo code of the CTAR function} %
\label{alg:ctar_function}
\textbf{Input:} Predicted action token sequence $\mathbf{a} = \{a_u\}_{u=1}^U$; Ground-truth action token sequence $\tilde{\mathbf{a}} = \{\tilde{a}_v\}_{v=1}^V$ \\[-1.1em]
\begin{algorithmic}[1]
    \State $\text{is\_valid} \gets \Call{FormatCheck}{\mathbf{a}}$
    
    \If{$\text{is\_valid} = \text{False}$}
        \State $\text{CTAR} \gets 0$ \Comment{Invalid format yields zero reward}
    \Else        
        \State $\mathbf{y} \triangleq (\mathbf{y}^{\text{pose}}, \mathbf{y}^{\text{grip}}) \gets \Call{Decode}{\mathbf{a}}$
        \State $\tilde{\mathbf{y}} \triangleq (\tilde{\mathbf{y}}^{\text{pose}}, \tilde{\mathbf{y}}^{\text{grip}}) \gets \Call{Decode}{\tilde{\mathbf{a}}}$
        \State $H \gets \text{Length}(\tilde{\mathbf{y}})$
        \State $R_{\text{sum}} \gets 0$
        \For{$t = 1$ \textbf{to} $H$}
            \vspace{0.1em} 
            \State $d_t \gets \dfrac{1}{\text{dim}(\mathbf{y}_t^{\text{pose}})} \| \mathbf{y}_t^{\text{pose}} - \tilde{\mathbf{y}}_t^{\text{pose}} \|_1$
            \vspace{0.15em} 
            \State $r_t^{\text{pose}} \gets \exp(-\alpha \cdot d_t)$
            \vspace{0.1em} 
            \State $r_t^{\text{grip}} \gets \mathbb{I}(\mathbf{y}_t^{\text{grip}} = \tilde{\mathbf{y}}_t^{\text{grip}})$
            \vspace{0.1em} 
            \State $r_t \gets \beta \cdot r_t^{\text{pose}} + (1 - \beta) \cdot r_t^{\text{grip}}$
            \vspace{0.1em} 
            \State $R_{\text{sum}} \gets R_{\text{sum}} + r_t$
        \EndFor
        \State $\text{CTAR} \gets R_{\text{sum}} / H$
        \EndIf
    \State \textbf{return} \text{CTAR}
\end{algorithmic}
\textbf{Output:} The CTAR score $\in [0, 1]$
\end{algorithm}
%

%
\begin{table*}[tp]
\centering
\caption{Multi-Task learning performance on SimplerEnv.}
\vspace{-0.1cm}
\label{tab:Simpler_multi_task_learning}
\tabcolsep=5pt
\normalsize
\resizebox{0.95\linewidth}{!}{%
\begin{tabular}{lc| ccccc cccc}
\toprule\toprule
\multirow{3}{*}{\textbf{Method}} & \multirow{3}{*}{\textbf{\shortstack{Training\\Strategy}}} & \multicolumn{5}{c}{\textbf{WidowX (Visual Matching)}} & \multicolumn{4}{c}{\textbf{Google Robot (Visual Matching)}} \\
\cmidrule(lr){3-7} \cmidrule(lr){8-11}
 & & \textbf{Put Carrot} & \textbf{Stack} & \textbf{Put Spoon} & \textbf{Put Eggplant} & \multirow{2}{*}{\textbf{Avg}} & \textbf{Pick Coke} & \textbf{Move} & \textbf{Open/Close} & \multirow{2}{*}{\textbf{Avg}} \\
 & & \textbf{on Plate} & \textbf{Blocks} & \textbf{on Towel} & \textbf{in Basket} & & \textbf{Can} & \textbf{Near} & \textbf{Drawer} & \\
\midrule

\rowcolor{gray!20} \multicolumn{11}{l}{\textit{\textbf{Continuous Action Models}}} \\
Octo-Base~\cite{team2024octo} & SFT & 8.3 & 0.0 & 12.5 & 43.1 & 16.0 & 17.0 & 4.2 & 22.7 & 16.8 \\
RoboVLM~\cite{liu2025towards}  & SFT & 25.0 & 12.5 & 29.2 & 58.3 & 31.3 & 77.3 & 61.7 & 43.5 & 63.4 \\
GR00T N1.5~\cite{nvidia2025gr00tn15} & SFT & -- & -- & -- & -- & -- & 69.3 & 68.7 & 35.8 & 52.4 \\
$\pi_0$~\cite{black2024pi_0} & SFT & \textbf{58.8} & 21.3 & 63.3 & {79.2} & 55.7 & 72.7 & 65.3 & 38.3 & 58.7 \\
ThinkAct~\cite{huang2025thinkact} & SFT + RFT & 37.5 & 8.7 & 58.3 & 70.8 & 43.8 & 92.0 & 72.4 & 50.0 & 71.5 \\
NORA-1.5~\cite{hung2025nora} & SFT & -- & -- & -- & -- & -- & 92.8 & 78.7 & 62.2 & 77.9 \\
NORA-1.5~\cite{hung2025nora} (DPO) & SFT+RFT & -- & -- & -- & -- & -- & 94.0 & \textbf{88.0} & \textbf{66.4} & \textbf{82.8} \\

\rowcolor{gray!20} \multicolumn{11}{l}{\textit{\textbf{Discrete Action Models}}} \\
TraceVLA~\cite{zheng2024tracevla} & SFT & -- & -- & -- & -- & -- & 28.0 & 53.7 & 57.0 & 42.0  \\
RT-1-X~\cite{brohan2022rt} & SFT & 4.2 & 0.0 & 0.0 & 0.0 & 1.1 & 56.7 & 31.7 & 59.7 & 53.4 \\
OpenVLA~\cite{kim2024openvla} & SFT & 0.0 & 0.0 & 0.0 & 4.1 & 1.0 & 16.3 & 46.2 & 35.6 & 27.7 \\
SpatialVLA~\cite{qu2025spatialvla} & SFT & 25.0 & 29.2 & 16.7 & \textbf{100.0} & 42.7 & 86.0 & 77.9 & 57.4 & 73.7 \\
$\pi_0$-FAST~\cite{pertsch2025fast} & SFT & 22.0 & \textbf{83.0} & 29.0 & 48.0 & 45.5 & 75.3 & 67.5 & 42.6 & 61.9 \\
NORA-1.5-FAST~\cite{hung2025nora} & SFT & -- & -- & -- & -- & -- & 88.6 & 86.4 & 41.2 & 72.1 \\
NORA-Long~\cite{hung2025nora_1} (Baseline) & SFT & 46.0 & 60.3 & 80.2 & 75.7 & 65.5 & 86.0 & 82.3 & 56.0 & 74.7 \\

\midrule
\rowcolor[HTML]{ECF4FF}
\textbf{NORA-Long}~\cite{hung2025nora_1} & \textbf{RFT (Ours)} & 50.2 & 64.4 & \textbf{84.3} & 77.0 & \textbf{69.0} & \textbf{94.0} &  84.7 & 58.5 & 79.1 \\ 
\addlinespace[0.8ex]
\rowcolor{blue!5}
$\boldsymbol{\Delta}$ & -- & \textcolor{blue}{+4.2} & \textcolor{blue}{+4.1} & \textcolor{blue}{+4.1} & \textcolor{blue}{+1.3} & \textcolor{blue}{\textbf{+3.5}} & \textcolor{blue}{+8.0} & \textcolor{blue}{+2.4} & \textcolor{blue}{+2.5} & \textcolor{blue}{+\textbf{4.4}} \\
\bottomrule\bottomrule
\end{tabular}%
}
\vspace{-0.2cm}
\end{table*}
\subsubsection{Format Compliance Reward}
%
%
While QACR and CTAR focus on optimizing prediction accuracy and control precision, their effectiveness is dependent on the structural validity of the generated outputs.
Specifically, the predicted sequences must adhere to the specified action dimensions and action chunk size.
To this end, we propose the Format Compliance Reward (FCR) to guide the model in generating structurally well-formed token sequences.
Concretely, we employ the Fast+ tokenizer to verify the compliance of the generated token sequence with the required output shape.
Accordingly, we define the FCR as a binary reward function that returns 1 if the validation passes and 0 otherwise.
The specific formulation is defined as:
\begin{equation}
  \text{FCR} = \begin{cases} 
    1, & \text{if valid} \\ 
    0, & \text{otherwise} 
  \end{cases}
  \label{eq:fcr_reward}
\end{equation}
where the condition ``valid'' indicates that the model output adheres to the predefined output format, enabling the Fast+ tokenizer to decode it into a continuous action chunk.
By explicitly incentivizing the model to acquire structurally valid output patterns, this reward establishes the necessary prerequisites for effective trajectory exploration.
Finally, we synthesize QACR, CTAR, and FCR to formulate the multi-dimensional process reward as follows:
\begin{equation}
  r = \omega \cdot \text{QACR} + (1 - \omega) \cdot \text{CTAR} + \lambda \cdot \text{FCR},
\end{equation}
where $\omega \in [0, 1]$ governs the trade-off between discrete action consistency and continuous trajectory alignment, and $\lambda$ scales the significance of structural format compliance.
\section{EXPERIMENTS}
In this section, we investigate the performance of LifeLong-RFT through comprehensive experiments in both simulation and real-world settings.
We first present the implementation details of the method, then detail the experimental configurations and results for both multi-task and continual learning.

\subsection{Implementation Details}
In our experiments, we adopt NORA-Long~\cite{hung2025nora_1} as the base VLA model, which utilizes the Fast+~\cite{pertsch2025fast} tokenizer for action representation.
During the reinforcement fine-tuning phase, the model undergoes full-parameter optimization.
Specifically, we set the rollout group size for GRPO to 8 and employ the AdamW~\cite{loshchilov2017decoupled} optimizer with a learning rate of $1 \times 10^{-6}$.
For the CTAR configuration, the hyperparameters $\alpha$ and $\beta$ are set to 5 and 0.8, respectively.
%
Finally, the multi-dimensional process reward is formulated as a weighted combination of rewards across three dimensions, with weighting coefficients $\omega=0.7$ and $\lambda=0.1$.
All experiments are conducted on 8 NVIDIA H20 GPUs.
%
%

\subsection{Multi-Task Learning Experiments}

\subsubsection{Experimental Setup}

\begin{figure*}[tp]
    \centering
    \includegraphics[width=\linewidth]{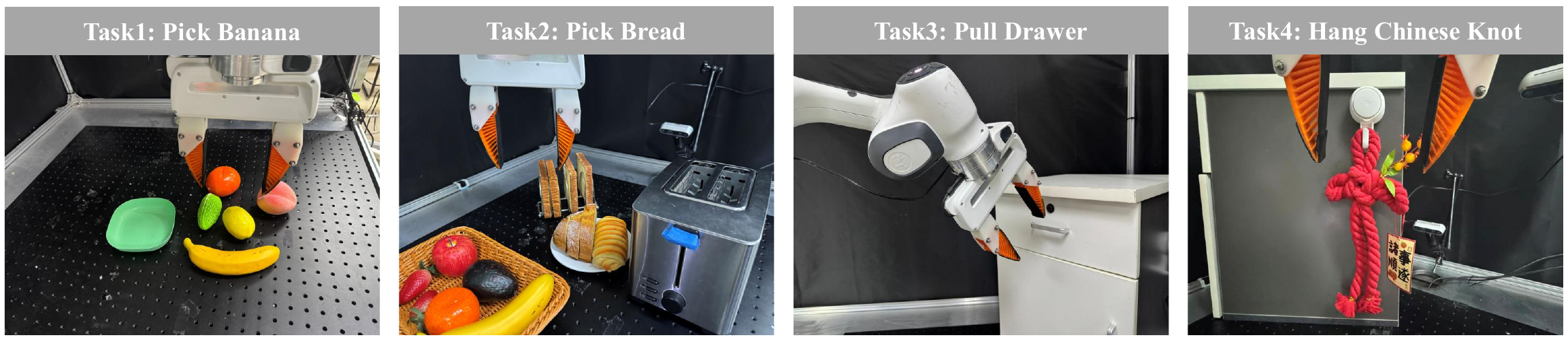}
    \captionof{figure}{
    \textbf{Overview of real-world experimental tasks:} Pick \& Place (Banana, Bread), Pull Drawer, and Hang Chinese Knot.
    }
\label{fig:realtask}
\vspace{-0.3cm}
\end{figure*}

\paragraph{Training Settings}
To evaluate multi-task learning in simulation, we utilize SimplerEnv~\cite{li2024evaluating} and LIBERO~\cite{liu2023libero}.
%
For SimplerEnv, we train the model on BridgeData V2~\cite{walke2023bridgedata} for WidowX and Fractal~\cite{brohan2022rt} for Google Robot.
%
For LIBERO, we fine-tune the model for each task suite (\ie, Object, Spatial, Goal, and Long), utilizing all 10 tasks with third-person and wrist inputs.
Moreover, we conduct experiments within real-world environments on the Franka robot, as shown in Fig.~\ref{fig:realtask}.
Concretely, we jointly train the model using 40 demonstrations each for the first three tasks, and 50 for the last.
\paragraph{Evaluation Protocols}
In SimplerEnv, we evaluate the model's performance on the WidowX and Google Robot platforms under the Visual Matching setting.
%
%
%
To ensure robust evaluation, each task is repeated over 24 trials under diverse initial object poses and environmental configurations.
For LIBERO, we evaluate the model on each task suite with 500 trials.
%
Additionally, in real-world experiments, we perform 20 trials per task.
Across all the above experiments, we report the average success rate (SR) as the evaluation metric.
\begin{table}[tp]
\centering
\caption{Multi-Task learning performance on LIBERO.}
\vspace{-0.1cm}
\label{tab:LIBERO_multi_task_learning}
\Large
\setlength{\tabcolsep}{3pt}
\resizebox{\columnwidth}{!}{%
\begin{tabular}{lc|cccc|c}
\toprule\toprule
\multirow{2}{*}[-3pt]{\textbf{Method}} & 
\multirow{2}{*}[-3pt]{\begin{tabular}{@{}c@{}}\textbf{Training}\\\textbf{Strategy}\end{tabular}} & 
\multicolumn{4}{c|}{\textbf{LIBERO}} & 
\multirow{2}{*}[-3pt]{\textbf{Avg}} \\
\cmidrule(lr){3-6}
 & & \textbf{Object} & \textbf{Spatial} & \textbf{Goal} & \textbf{Long} & \\
\midrule
\multicolumn{7}{l}{\cellcolor{gray!20}\textit{\textbf{Continuous Action Models}}} \\
\midrule
Octo-Base~\cite{team2024octo} & SFT & 85.7 & 78.9 & 84.6 & 51.1 & 75.1  \\
GR00T N1~\cite{bjorck2025gr00t} & SFT & 97.6 & 94.4 & 93.0 & 90.6 & 93.9 \\
$\pi_0$~\cite{black2024pi_0}  & SFT & 98.8 & 96.8 & 95.8 & 85.2 & 94.2 \\
OpenVLA-OFT~\cite{kim2025fine} & SFT & 98.1 & 96.9 & 95.5 & \textbf{91.1} & 95.4 \\
ThinkAct~\cite{huang2025thinkact} & SFT + RFT & 91.4 & 88.3 & 87.1 & 70.9 & 84.4 \\
VLA-RFT~\cite{li2025vla} & SFT + RFT & 94.4 & 94.4 & 95.4 & 80.2 & 91.1 \\
NORA-1.5~\cite{hung2025nora} & SFT & 96.4 & 97.3 & 94.5 & 89.6 & 94.5 \\
NORA-1.5~\cite{hung2025nora} (DPO) & SFT + RFT & 96.0 & 98.0 & 95.4 & 90.5 & 95.0 \\
\midrule
\multicolumn{7}{l}{\cellcolor{gray!20}\textit{\textbf{Discrete Action Models}}} \\
\midrule
TraceVLA~\cite{zheng2024tracevla} & SFT & 85.2 & 84.6 & 75.1 & 54.1 & 74.8 \\
OpenVLA~\cite{kim2024openvla} & SFT & 88.4 & 84.7 & 79.2 & 53.7 & 76.5 \\
SpatialVLA~\cite{qu2025spatialvla} & SFT & 89.9 & 88.2 & 78.6 & 55.5 & 78.1 \\
CoT-VLA~\cite{zhao2025cot} & SFT & 91.6 & 87.5 & 87.6 & 69.0 & 83.9 \\
WorldVLA~\cite{cen2025worldvla} & SFT & 96.2 & 87.6 & 83.4 & 60.0 & 79.1 \\
$\pi_{0}$-Fast~\cite{pertsch2025fast} & SFT & 96.8 & 96.4 & 88.6 & 60.2 & 85.5 \\
MolmoAct-7B-D~\cite{lee2025molmoact} & SFT & 95.4 & 87.0 & 87.6 & 77.2 & 86.6 \\
TGRPO~\cite{chen2025tgrpo} & SFT + RFT & 92.2 & 90.4 & 81.0 & 59.2 & 80.7 \\

NORA-Long~\cite{hung2025nora_1} (Baseline) & SFT & 97.5 & 96.4 & 91.0 & 82.4 & 91.8 \\

\midrule 

\rowcolor[HTML]{ECF4FF}
\textbf{NORA-Long}~\cite{hung2025nora_1} & \textbf{RFT (Ours)} & \textbf{99.2} & \textbf{98.2} & \textbf{95.8} & 89.0 & \textbf{95.6} \\

\addlinespace[0.8ex]

\rowcolor{blue!5}
\textbf{$\boldsymbol{\Delta}$} & -- & \textcolor{blue}{+1.7} & \textcolor{blue}{+1.8} & \textcolor{blue}{+4.8} & \textcolor{blue}{+6.6} & \textcolor{blue}{\textbf{+3.8}} \\

\bottomrule\bottomrule
\end{tabular}%
}
\vspace{-0.3cm}
\end{table}
\subsubsection{Performance on Simulation}
%
%
In Table~\ref{tab:Simpler_multi_task_learning}, LifeLong-RFT consistently improves the performance of the SFT baseline across diverse evaluation scenarios, achieving average success rate improvements of 3.5\% on WidowX and 4.4\% on the Google Robot.
%
%
Moreover, results in Table~\ref{tab:LIBERO_multi_task_learning} show that our method surpasses all competing continuous and discrete action models, achieving a superior average success rate of 95.6\%.

\subsubsection{Performance on Real-World}
Beyond simulation, we also conducted real-world experiments.
Table~\ref{tab:real_world_multi_task_learning} demonstrates that our method consistently outperforms all competing methods across four tasks.
Specifically, when employing NORA-Long as the backbone, LifeLong-RFT achieves an average success rate improvement of 8.7\% over the SFT baseline.
Notably, for the dexterous task ``Hang Chinese Knot", this method outperforms the SFT baseline by 15\%.
%
%
\subsection{Continual Learning Experiments}

\subsubsection{Experimental Setup}
\paragraph{Training Settings}
%
%
We utilize LIBERO~\cite{liu2023libero} to conduct experiments within simulated environments.
Following LOTUS~\cite{wan2024lotus}, the training process consists of a base task stage and a lifelong learning stage.
%
%
%
For each task suite, we conduct the base task stage training using its first six tasks, each comprising 50 demonstrations.
Subsequently, the lifelong learning stage focuses on incremental learning for the remaining four tasks.
In this stage, each new task consists of only 10 demonstrations, while 5 demonstrations per previously learned task are retained for Experience Replay (ER)~\cite{chaudhry2019tiny, li2024generalizing}.
Overall, a complete experimental cycle comprises one base learning step and four sequential lifelong learning steps.
Additionally, for real-world experiments, we sequentially train the model on four tasks as illustrated in Fig.~\ref{fig:realtask}, utilizing 20 demonstrations for each new task and retaining 5 for each previous task.
%
%
%
\begin{table}[tp]
\centering
\caption{Multi-Task learning performance on real-world.}
\vspace{-0.1cm}
\label{tab:real_world_multi_task_learning}
\tabcolsep=5pt
\normalsize
\renewcommand{\arraystretch}{1.0}
\renewcommand\theadgape{}
\newcommand{\pdm}[1]{{\color{black!60}\normalsize #1}}
\resizebox{\columnwidth}{!}{%
\begin{tabular}{l | cc | cc >{\columncolor{blue!5}[2pt][\tabcolsep]}c}
\toprule\toprule
\multirow{2}{*}{\textbf{Task Split}} & \thead{$\boldsymbol{\pi_0}$}~\cite{black2024pi_0} & \thead{\textbf{OpenVLA}}~\cite{kim2024openvla} & \multicolumn{3}{c}{\textbf{NORA-Long}~\cite{hung2025nora}} \\
\cmidrule(lr){2-2} \cmidrule(lr){3-3} \cmidrule(lr){4-6}
 & \pdm{SFT} & \pdm{SFT} & \pdm{SFT} & \textbf{RFT (Ours)} & \multicolumn{1}{c}{\textbf{$\boldsymbol\Delta$}} \\
\midrule
Pick Banana       & 90.0 & 75.0 & 85.0 & 90.0 & \textcolor{blue}{+5.0} \\
Pick Bread        & 75.0 & 70.0 & 75.0 & 85.0 & \textcolor{blue}{+10.0} \\
Pull Drawer       & 95.0 & 85.0 & 95.0 & 100.0 & \textcolor{blue}{+5.0} \\
Hang Chinese Knot & 65.0 & 55.0 & 60.0 & 75.0 & \textcolor{blue}{+15.0} \\
\rowcolor[HTML]{ECF4FF}
\textbf{Overall}  & 81.3 & 71.3 & 78.8 & \textbf{87.5} & \cellcolor{blue!5}\textcolor{blue}{\textbf{+8.7}} \\
\bottomrule\bottomrule
\end{tabular}
}
\vspace{-0.3cm}
\end{table}
\begin{table*}[tp]
\centering
\caption{Continual learning performance on LIBERO.}
\vspace{-0.1cm}
\label{tab:simulation_continual_learning}
\tabcolsep=5pt
\normalsize
\renewcommand{\arraystretch}{1.}
\renewcommand\theadgape{}
\newcommand{\pdm}[1]{{\color{black!60}\normalsize #1}}
\definecolor{brickred}{HTML}{800000}

\resizebox{0.95\linewidth}{!}{%
\begin{tabular}{ll|cccccc|cc >{\columncolor{blue!5}[2pt][\tabcolsep]}c}
\toprule\toprule
\multirow{2}{*}{\textbf{Task Split}} & \multirow{2}{*}{\textbf{Metrics}} & \thead{\textbf{BUDS}}~\cite{zhu2022bottom} & \thead{\textbf{LOTUS}}~\cite{wan2024lotus} & \thead{\textbf{SPECI}}~\cite{xu2025speci} & \thead{$\boldsymbol{\pi_0}$}~\cite{black2024pi_0} & \thead{\textbf{OpenVLA}}~\cite{kim2024openvla} & \thead{\textbf{OpenVLA-OFT}}~\cite{kim2025fine} & \multicolumn{3}{c}{\textbf{NORA-Long}~\cite{hung2025nora_1}} \\
\cmidrule(lr){3-3} \cmidrule(lr){4-4} \cmidrule(lr){5-5} \cmidrule(lr){6-6} \cmidrule(lr){7-7} \cmidrule(lr){8-8} \cmidrule(lr){9-11}
& & \pdm{BC} & \pdm{BC} & \pdm{BC} & \pdm{SFT} & \pdm{SFT} & \pdm{SFT} & \pdm{SFT} & \textbf{RFT (Ours)} & \multicolumn{1}{c}{\thead{$\boldsymbol{\Delta}$}} \\
\midrule

\multirow{3}{*}{LIBERO-Object}
 & FWT ($\uparrow$) & 52.0 & 74.0 & 83.0 & 73.0 & 59.4 & 89.8 & 84.8 & 96.0 & \textcolor{blue}{+11.2} \\
 & NBT ($\downarrow$) & 21.0 & 11.0 & 10.0 & 16.2 & 17.9 & 3.1 & 6.8 & 1.5 & \textcolor{brickred}{-5.3} \\
 \rowcolor[HTML]{ECF4FF}
 & \textbf{AUC} ($\uparrow$) & 47.0 & 65.0 & 78.0 & 59.3 & 45.1 & 87.4 & 79.7 & \textbf{94.8} & \cellcolor{blue!5}\textcolor{blue}{\textbf{+15.1}} \\
\midrule

\multirow{3}{*}{LIBERO-Spatial}
 & FWT ($\uparrow$) & -- & -- & 67.0 & 74.4 & 64.2 & 88.6 & 82.8 & 94.0 & \textcolor{blue}{+11.2} \\
 & NBT ($\downarrow$) & -- & -- & 6.0 & 23.7 & 17.6 & 9.4 & 14.0 & 3.7 & \textcolor{brickred}{-10.3} \\
 \rowcolor[HTML]{ECF4FF}
 & \textbf{AUC} ($\uparrow$) & -- & -- & 66.0 & 55.5 & 50.8 & 81.7 & 71.7 & \textbf{91.2} & \cellcolor{blue!5}\textcolor{blue}{\textbf{+19.5}} \\
\midrule

 & FWT ($\uparrow$) & 50.0 & 61.0 & 74.0 & 74.6 & 58.6 & 90.2 & 72.8 & 92.4 & \textcolor{blue}{+19.6} \\
\multirow{-2}{*}{LIBERO-Goal}
 & NBT ($\downarrow$) & 39.0 & 30.0 & 20.0 & 23.9 & 5.8 & 13.8 & 25.2 & 3.1 & \textcolor{brickred}{-22.1} \\
 \rowcolor[HTML]{ECF4FF}
 & \textbf{AUC} ($\uparrow$) & 42.0 & 56.0 & 65.0 & 56.3 & 53.5 & 79.2 & 54.4 & \textbf{90.3} & \cellcolor{blue!5}\textcolor{blue}{\textbf{+35.9}} \\
\midrule

\multirow{3}{*}{LIBERO-Long}
 & FWT ($\uparrow$) & -- & -- & 58.0 & 53.8 & 32.0 & 64.0 & 61.0 & 74.2 & \textcolor{blue}{+13.2} \\
 & NBT ($\downarrow$) & -- & -- & 21.0 & 14.2 & 14.1 & 31.4 & 17.3 & 12.8 & \textcolor{brickred}{-4.5} \\
 \rowcolor[HTML]{ECF4FF}
 & \textbf{AUC} ($\uparrow$) & -- & -- & 46.0 & 42.5 & 20.8 & 38.7 & 47.3 & \textbf{64.5} & \cellcolor{blue!5}\textcolor{blue}{\textbf{+17.2}} \\
\bottomrule\bottomrule
\end{tabular}
}
\vspace{-0.3cm}
\end{table*}
\paragraph{Evaluation Protocols}
We utilize three metrics~\cite{liu2023libero} to assess the model's continual learning capabilities: Forward Transfer (FWT), Negative Backward Transfer (NBT), and Area Under the Success Rate Curve (AUC).
All three metrics are derived from the task success rate.
Specifically, a higher FWT indicates improved adaptation to new tasks; a lower NBT implies effective mitigation of catastrophic forgetting of previously learned tasks; and a higher AUC reflects better average success rates across all evaluated tasks.
Given that the model sequentially learns over $K$ tasks $\{\mathcal{T}_k\}_{k=1}^{K}$, let $s_{k,j}$ denote the agent's success rate on task $j$ after learning the first $k$ tasks.
These three metrics are defined as follows:
$\text{FWT} = \sum_{k \in [K]} \frac{s_{k,k}}{K}$,
$\text{NBT} = \sum_{k \in [K]} \frac{\text{NBT}_k}{K}$,
$\text{NBT}_k = \frac{1}{K-k} \sum_{q=k+1}^{K} (s_{k,k} - s_{q,k})$, and
$\text{AUC} = \sum_{k \in [K]} \frac{\text{AUC}_k}{K}$,
$\text{AUC}_k = \frac{1}{K-k+1} (s_{k,k} + \sum_{q=k+1}^{K} s_{q,k})$.
In our experiments, we evaluate policies on all learned tasks, conducting 50 episodes for LIBERO and 20 episodes for real-world experiments.
%
%

%
\begin{table}[tp]
\centering
\caption{Continual learning performance on real-world.}
\vspace{-0.1cm}
\label{tab:real_world_continual_learning}
\tabcolsep=6pt
\Large
\renewcommand{\arraystretch}{1.0}
\renewcommand\theadgape{}

\newcommand{\pdm}[1]{{\color{black!60}\Large #1}}
\definecolor{brickred}{HTML}{800000}

\resizebox{\columnwidth}{!}{
\begin{tabular}{ll|cc|cc >{\columncolor{blue!5}[2pt][\tabcolsep]}c}
\toprule\toprule
\multirow{2}{*}{\textbf{Task Split}} & \multirow{2}{*}{\textbf{Metrics}} & \thead{$\boldsymbol{\pi_0}$}~\cite{black2024pi_0} &  \thead{\textbf{OpenVLA}}~\cite{kim2024openvla}  & \multicolumn{3}{c}{\textbf{NORA-Long}~\cite{hung2025nora_1}} \\
\cmidrule(lr){3-3} \cmidrule(lr){4-4} \cmidrule(lr){5-7}
& & \pdm{SFT} & \pdm{SFT} & \pdm{SFT} & \textbf{RFT (Ours)} & \multicolumn{1}{c}{\thead{$\boldsymbol{\Delta}$}} \\
\midrule

\multirow{3}{*}{Real-World}
& FWT ($\uparrow$) & 58.8 & 46.3 & 56.3 & 80.0 & \textcolor{blue}{+23.7} \\
& NBT ($\downarrow$) & 16.3 & 17.8 & 18.3 & 6.1 & \textcolor{brickred}{-12.2} \\
\rowcolor[HTML]{ECF4FF}
& \textbf{AUC} ($\uparrow$) & 47.9 & 35.1 & 44.2 & \textbf{75.9} & \cellcolor{blue!5}\textcolor{blue}{\textbf{+31.7}} \\
\bottomrule\bottomrule
\end{tabular}
}
\vspace{-0.cm}
\end{table}

\subsubsection{Performance on Simulation}
%
First, we evaluate LifeLong-RFT on LIBERO to validate its continual learning capabilities.
We compare against models~\cite{zhu2022bottom,wan2024lotus,xu2025speci} trained with behavioral cloning (BC) loss.
Additionally, we assess large-scale VLAs~\cite{black2024pi_0,kim2024openvla,kim2025fine,hung2025nora_1} optimized by SFT.
As shown in Table~\ref{tab:simulation_continual_learning}, our method consistently outperforms other methods across all task suites.
Notably, on LIBERO-Goal, LifeLong-RFT demonstrates significant superiority, achieving a substantial gain of 35.9 in AUC over the SFT baseline.

\subsubsection{Performance on Real-World}
%
Furthermore, we assess real-world continual learning performance.
As presented in Table~\ref{tab:real_world_continual_learning}, our method shows a substantial improvement of 23.7 in FWT over the SFT baseline, significantly outperforming the other two models.
%
%
Furthermore, the approach yields an NBT of only 6.1, demonstrating a strong capability to preserve performance on learned tasks.
Overall, the model fine-tuned with LifeLong-RFT achieves an average success rate of 75.9\% across the learning cycle, exhibiting robust continual learning.

\subsection{Ablation Studies}

\subsubsection{Effectiveness of Multi-Dimensional Process Reward}
To verify the effectiveness of multi-dimensional process reward, we conduct multi-task learning experiments on LIBERO.
The first row of Table~\ref{tab:ablation_of_MDPR} indicates that removing QACR leads to a 2.8\% average performance drop across the four task suites, confirming its necessity for accurate quantized action prediction.
%
%
%
The second row further underscores the critical role of CTAR, where its exclusion leads to a 90.9\% performance drop, resulting in the model being nearly incapable of task completion.
%
%
Additionally, the third row shows that FCR is critical for guaranteeing the structural validity of the output.
Particularly in LIBERO-Long, the absence of FCR leads to a 4.4\% degradation.
Overall, each reward component exhibits consistent effectiveness, enhancing overall performance.
\begin{table}[tp]
\centering
\caption{Ablation of the multi-dimensional process reward.}
\vspace{-0.1cm}
\label{tab:ablation_of_MDPR}

\definecolor{brickred}{HTML}{800000} 

\newcommand{\bad}[1]{\cellcolor{blue!5}\textcolor{brickred}{#1}}
\renewcommand{\arraystretch}{1.1}
\renewcommand\theadgape{}
\tabcolsep=5pt 
\Large
\resizebox{\columnwidth}{!}{%
\begin{tabular}{l | cc | cc | cc | cc | >{\columncolor{gray!5}}cc}
\toprule\toprule
\multirow{2}{*}{\textbf{Settings}} & \multicolumn{2}{c|}{\textbf{Object}} & \multicolumn{2}{c|}{\textbf{Spatial}} & \multicolumn{2}{c|}{\textbf{Goal}} & \multicolumn{2}{c|}{\textbf{Long}} & \multicolumn{2}{c}{\textbf{Avg}} \\
\cmidrule(lr){2-3} \cmidrule(lr){4-5} \cmidrule(lr){6-7} \cmidrule(lr){8-9} \cmidrule(lr){10-11}
& \textbf{SR} & $\boldsymbol\Delta$ & \textbf{SR} & $\boldsymbol\Delta$ & \textbf{SR} & $\boldsymbol\Delta$ & \textbf{SR} & $\boldsymbol\Delta$ & \textbf{SR} & $\boldsymbol\Delta$ \\
\midrule

w/o QACR & 97.0 & \bad{-2.2} & 96.4 & \bad{-1.8} & 92.2 & \bad{-3.6} & 85.6 & \bad{-3.4} & 92.8 & \bad{-2.8} \\
w/o CTAR & 8.0  & \bad{\textbf{-91.2}}& 6.2  & \bad{\textbf{-92.0}}& 2.4  & \bad{\textbf{-93.4}}& 2.0  & \bad{\textbf{-87.0}}& 4.7  & \bad{\textbf{-90.9}} \\
w/o FCR  & 98.0 & \bad{-1.2} & 96.2 & \bad{-2.0} & 93.2 & \bad{-2.6} & 84.6 & \bad{-4.4} & 93.0 & \bad{-2.6} \\

\midrule
 \rowcolor[HTML]{ECF4FF}
\textbf{RFT (Ours)} & \textbf{99.2} & \textbf{-} & \textbf{98.2} & \textbf{-} & \textbf{95.8} & \textbf{-} & \textbf{89.0} & \textbf{-} & \textbf{95.6} & \textbf{-} \\
\bottomrule\bottomrule
\end{tabular}
}
\vspace{-0.3cm}
\end{table}
\begin{figure}[tp]
    \centering
    \includegraphics[width=\linewidth]{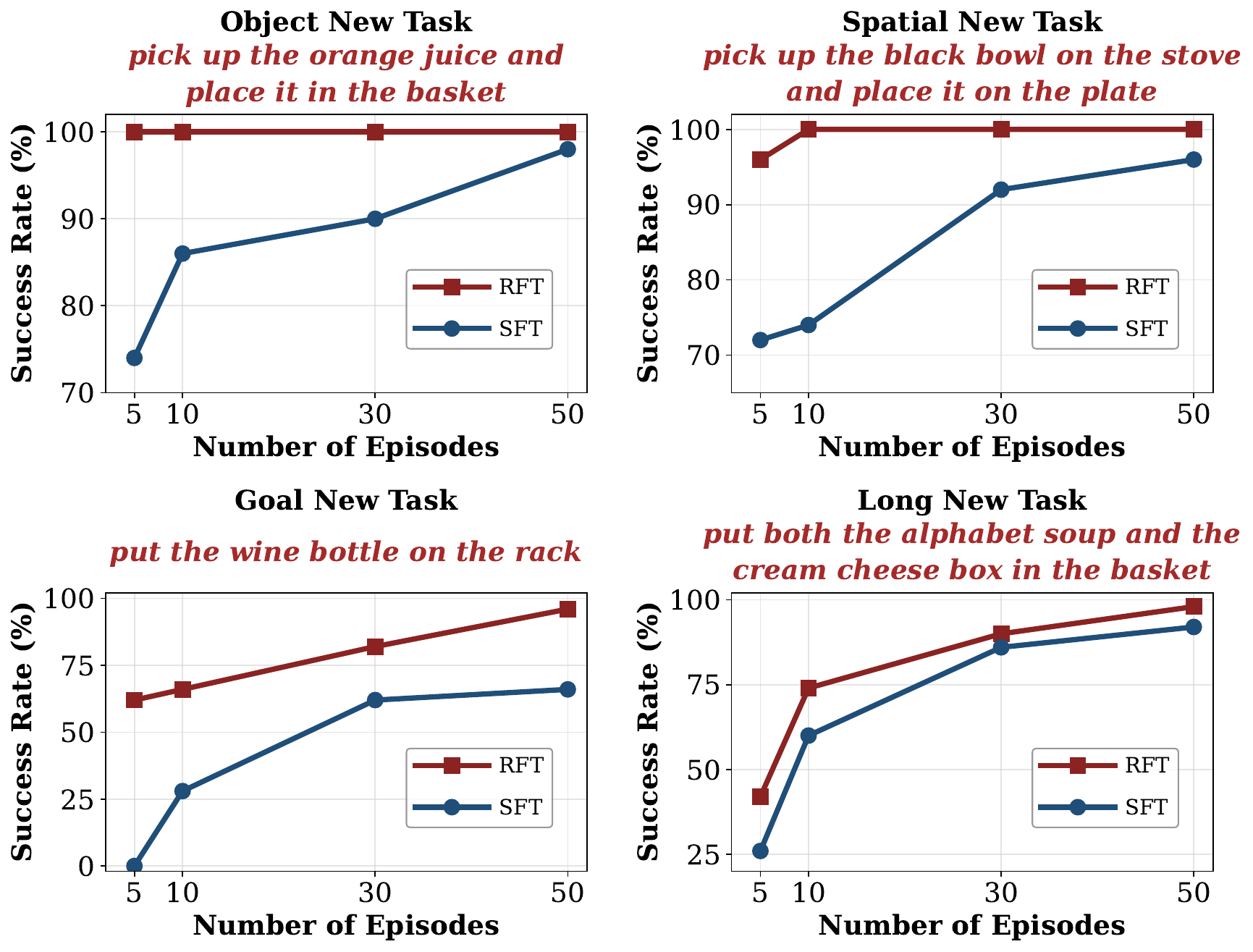}
    \captionof{figure}{Adaptation efficiency on representative new tasks.}
\label{fig:efficiency}
\vspace{-0.3cm}
\end{figure}

\subsubsection{Efficiency of New Task Adaptation}
To assess the adaptation efficiency of LifeLong-RFT in acquiring new tasks, we conduct experiments across the four task suites of the LIBERO benchmark.
Specifically, we first train the model on the initial six tasks of each suite via multi-task learning.
Subsequently, we select a representative task from the remaining four unseen tasks in each suite to evaluate the model's efficiency for new task adaptation.
%
%
In Fig.~\ref{fig:efficiency}, LifeLong-RFT shows superior adaptation efficiency compared to SFT.
On the ``Pick Orange Juice" task, it achieves 100\% success using merely 5 demonstrations, whereas SFT reaches 98\% even when trained with 50.
Similarly, for the ``Pick Bowl" task, LifeLong-RFT matches the performance of the SFT baseline trained on 50 demonstrations using only 5, and improves to 100\% success with just 10.
In addition to few-shot scenarios, our approach sustains its advantage with the full set of demonstrations: it surpasses SFT by 30\% on the ``Put Wine Bottle" task and exhibits superior performance on the long-horizon task ``Put Alphabet Soup and Cream Cheese".
%
%
\section{Conclusion and Future Outlook}
In this work, we introduce LifeLong-RFT, a reinforcement fine-tuning strategy to overcome the extensive data requirements and catastrophic forgetting associated with SFT.
%
%
Unlike existing methods, our method integrates the chunking-level on-policy RL with the multi-dimensional process reward mechanism to achieve efficient new task adaptation while preserving pre-existing knowledge. 
Specifically, this mechanism employs Quantized Action Consistency Reward, Continuous Trajectory Alignment Reward, and Format Compliance Reward to quantify the heterogeneous contributions of action chunks across three dimensions, independent of environmental feedback and pre-trained reward models.
Comprehensive experiments indicate that LifeLong-RFT consistently surpasses SFT-based methods in both multi-task and continual learning.
%

%
\noindent\textbf{Limitations \& Future Work.}
This work primarily focuses on discrete action models, yet their performance falls short of the levels achieved by continuous action models. Future research extending the LifeLong-RFT training strategy to continuous action models will significantly accelerate the transition of VLAs from laboratory research to industrial applications.
%

\section{ACKNOWLEDGMENTS}
%
%
This work was supported by the New Generation Artificial Intelligence-National Science and Technology Major Project 2025ZD0122902; in part by the National Natural Science Foundation of China (NSFC) under Grants 62136008 and 62293545; the Beijing Major Science and Technology Project under Contract no. Z251100008125023; the Suzhou Innovation and Entrepreneurship Leading Talents Programme – Innovation Leading Talent in Universities and Research Institutes under Grant ZXL2025310; and the Beijing Academy of Artificial Intelligence (BAAI).
%


\bibliographystyle{plainnat}
\bibliography{references}
\clearpage
\maketitlesupplementary
\appendix
\setcounter{page}{1}

\subsection{Training Details}
In this section, we detail the training settings for multi-task learning and continual learning in both simulation (\ie, SimplerEnv~\cite{li2024evaluating} and LIBERO~\cite{liu2023libero}) and real-world environments.

\subsubsection{Multi-Task Learning}
The training settings for multi-task learning on SimplerEnv are detailed in Table~\ref{tab:multi_task_setting_Simpler}.
Notably, the WidowX setup utilizes a global batch size of 512 for 30 epochs, whereas the Google Robot employs a batch size of 1024 for 40 epochs.
Apart from these specific adjustments, the remaining hyperparameters are kept consistent, highlighting the cross-platform robustness of our approach.
\begin{table}[ht]
    \centering
    \caption{Multi-Task learning settings on SimplerEnv.}
    \label{tab:multi_task_setting_Simpler}
    \small
    \setlength{\tabcolsep}{10pt} 
    \renewcommand{\arraystretch}{0.9} 
    \begin{tabular}{lcc}
        \toprule\toprule
        \textbf{Hyperparameter} & \textbf{WidowX} & \textbf{Google Robot} \\ 
        \midrule
        
        \multicolumn{3}{l}{\cellcolor{gray!20}\textit{\textbf{Platform-Specific Settings}}} \\
        \midrule
        Global Batch Size              & 512 & 1024 \\
        Epochs                  & 30 & 40  \\
        \midrule
        
        \multicolumn{3}{l}{\cellcolor{gray!20}\textit{\textbf{Shared Settings}}} \\
        \midrule
        Learning Rate           & \multicolumn{2}{c}{$1 \times 10^{-6}$} \\
        Optimizer               & \multicolumn{2}{c}{AdamW~\cite{loshchilov2017decoupled}} \\
        Group Size              & \multicolumn{2}{c}{8} \\
        Temperature             & \multicolumn{2}{c}{0.8} \\
        $(\alpha,\beta,\omega,\lambda, \gamma)$ & \multicolumn{2}{c}{$(5, 0.8, 0.7, 0.1, 0.001)$} \\
        \bottomrule\bottomrule
    \end{tabular}
\end{table}
Table~\ref{tab:multi_task_setting_LIBERO} details the hyperparameter settings for multi-task learning on LIBERO.
Specifically, for the long-horizon task suite LIBERO-Long, we set the global batch size to 256 and train for 35 epochs.
The remaining three task suites share a unified parameter configuration with 15 training epochs.
\begin{table}[ht]
    \centering
    \caption{Multi-Task learning settings on LIBERO.}
    \label{tab:multi_task_setting_LIBERO}
    \small
    \setlength{\tabcolsep}{6pt} 
    \renewcommand{\arraystretch}{0.9} 
    \begin{tabular}{lcc}
        \toprule\toprule
        \textbf{Hyperparameter} & \textbf{Object / Spatial / Goal} & \textbf{Long} \\ 
        \midrule
        \multicolumn{3}{l}{\cellcolor{gray!20}\textit{\textbf{Task-Specific Settings}}} \\
        \midrule
        Global Batch Size              & 128 & 256 \\
        Epochs                  & 15 & 35  \\
        \midrule
        \multicolumn{3}{l}{\cellcolor{gray!20}\textit{\textbf{Shared Settings}}} \\
        \midrule
        Learning Rate           & \multicolumn{2}{c}{$1 \times 10^{-6}$} \\
        Optimizer               & \multicolumn{2}{c}{AdamW~\cite{loshchilov2017decoupled}} \\
        Group Size              & \multicolumn{2}{c}{8} \\
        Temperature             & \multicolumn{2}{c}{0.8} \\
        $(\alpha,\beta,\omega,\lambda, \gamma)$ & \multicolumn{2}{c}{$(5, 0.8, 0.7, 0.1, 0.001)$} \\
        \bottomrule\bottomrule
    \end{tabular}
\end{table}

For the four real-world tasks on the Franka robot, totaling 170 demonstrations, the training parameters are provided in Table~\ref{tab:multi_task_training_setting_real_world}.
We set the global batch size to 128 and train for 20 epochs, while all other parameters remain consistent with the simulation experiments.

\begin{table}[ht]
    \centering
    \caption{Multi-Task learning settings on real-world tasks.}
    \label{tab:multi_task_training_setting_real_world}
    \small
    
    \setlength{\tabcolsep}{11pt}  
    \renewcommand{\arraystretch}{1.} 
    
    \begin{tabular}{lc} 
        \toprule\toprule
        \textbf{Hyperparameter} & \textbf{Real-World} \\ 
        \midrule
        \multicolumn{2}{l}{\cellcolor{gray!20}\textit{\textbf{Shared Settings}}} \\
        \midrule
        Global Batch Size                       & 128            \\
        Epochs                                  & 20             \\
        Learning Rate                           & $1 \times 10^{-6}$ \\
        Optimizer                               & AdamW~\cite{loshchilov2017decoupled}          \\
        Group Size                              & 8              \\
        Temperature                             & 0.8            \\
        $(\alpha,\beta,\omega,\lambda, \gamma)$ & $(5, 0.8, 0.7, 0.1, 0.001)$ \\
        \bottomrule\bottomrule
    \end{tabular}
\end{table}

\subsubsection{Continual Learning}

\textbf{(1)} The continual learning protocol in LIBERO consists of an initial base task stage and a subsequent lifelong learning stage.
For the base task stage, the training parameters remain consistent with Table~\ref{tab:multi_task_setting_LIBERO}, while the configurations for the four task suites in the lifelong learning stage are presented in Table~\ref{tab:continual_learning_setting_LIBERO_Real_World}.
Given that the lifelong learning stage utilizes limited demonstrations to learn new tasks, we set the global batch size to 32 and train for 10 epochs.
\textbf{(2)} The real-world continual learning experiment includes only the lifelong learning stage, requiring the model to learn four tasks sequentially.
As demonstrated in Table~\ref{tab:continual_learning_setting_LIBERO_Real_World}, the training configurations remain consistent with LIBERO.
\begin{table}[ht]
    \centering
    \caption{Continual learning settings for LIBERO and real-world experiments.}
    \label{tab:continual_learning_setting_LIBERO_Real_World}
    \small
    \setlength{\tabcolsep}{11pt}  
    \renewcommand{\arraystretch}{1.} 
    \begin{tabular}{lc}
        \toprule\toprule
        \textbf{Hyperparameter} & \textbf{LIBERO / Real-World} \\ 
        \midrule
        \multicolumn{2}{l}{\cellcolor{gray!20}\textit{\textbf{Shared Settings}}} \\
        \midrule
        Global Batch Size                       & 32             \\
        Epochs                                  & 10             \\
        Learning Rate                           & $1 \times 10^{-6}$ \\
        Optimizer                               & AdamW~\cite{loshchilov2017decoupled} \\
        Group Size                              & 8              \\
        Temperature                             & 0.8            \\
        $(\alpha,\beta,\omega,\lambda, \gamma)$ & $(5, 0.8, 0.7, 0.1, 0.001)$ \\
        \bottomrule\bottomrule
    \end{tabular}
\end{table}
%

%
%
%

%
%

\subsection{Additional Experimental Results and Analysis}

\subsubsection{Detailed Continual Learning Results}
To comprehensively analyze the continual learning effectiveness of LifeLong-RFT, we report detailed results for the model on all learned tasks at each training phase.
As shown in Table~\ref{tab:detailed_continual_learning_Libero_All}, our method exhibits strong performance in adapting to new tasks and retaining prior knowledge.
Notably, following the training of Task 8 in the LIBERO-Goal suite, the model exhibits performance improvements on previously learned tasks (\ie, Tasks 2, 3, and 7), demonstrating strong backward transfer capabilities.
However, within the long-horizon LIBERO-Long suite, it exhibits suboptimal performance on certain tasks (such as Task 7 at 36\% and Task 9 at 34\%) with limited demonstrations.
This underscores a challenge worthy of further exploration in future work.
\begin{table*}[tp]
    \centering
    \caption{Detailed continual learning results on four LIBERO task suites (Object, Spatial, Goal, and Long).}
    \label{tab:detailed_continual_learning_Libero_All}
    \vspace{-0.1cm}
    \Large
    \setlength{\tabcolsep}{5pt}
    \renewcommand{\arraystretch}{1.}
    \resizebox{\linewidth}{!}{%
        \begin{tabular}{l|cccccccccc|cccccccccc}
            \toprule\toprule
            \multirow{2}{*}{\textbf{Task Split}} & \multicolumn{10}{c|}{\textbf{LIBERO-Object}} & \multicolumn{10}{c}{\textbf{LIBERO-Spatial}} \\
            \cmidrule(lr){2-11} \cmidrule(lr){12-21}
             & \textbf{T-1} & \textbf{T-2} & \textbf{T-3} & \textbf{T-4} & \textbf{T-5} & \textbf{T-6} & \textbf{T-7} & \textbf{T-8} & \textbf{T-9} & \textbf{T-10} & \textbf{T-1} & \textbf{T-2} & \textbf{T-3} & \textbf{T-4} & \textbf{T-5} & \textbf{T-6} & \textbf{T-7} & \textbf{T-8} & \textbf{T-9} & \textbf{T-10} \\
            \midrule
            
            \multicolumn{21}{l}{\cellcolor{gray!20}\textit{\textbf{Base Task Stage}}} \\
            \midrule
            Base Task 1-6 & 100\% & 100\% & 100\% & 98\% & 98\% & 100\% & -- & -- & -- & -- & 90\% & 100\% & 98\% & 98\% & 96\% & 84\% & -- & -- & -- & -- \\
            \midrule
            
            \multicolumn{21}{l}{\cellcolor{gray!20}\textit{\textbf{LifeLong Learning Stage}}} \\
            \midrule
            New Task 7  & 92\% & 96\% & 98\% & 96\% & 98\% & 100\% & 96\% & -- & -- & -- & 94\% & 92\% & 98\% & 84\% & 94\% & 96\% & 100\% & -- & -- & -- \\
            New Task 8  & 98\% & 100\% & 94\% & 98\% & 96\% & 100\% & 100\% & 82\% & -- & -- & 100\% & 97\% & 100\% & 94\% & 86\% & 92\% & 98\% & 90\% & -- & -- \\
            New Task 9  & 96\% & 96\% & 96\% & 86\% & 96\% & 100\% & 98\% & 92\% & 96\% & -- & 70\% & 80\% & 98\% & 92\% & 92\% & 88\% & 96\% & 94\% & 90\% & -- \\
            \rowcolor[HTML]{ECF4FF} 
            \textbf{New Task 10} & 94\% & 100\% & 100\% & 96\% & 96\% & 94\% & 100\% & 76\% & 92\% & 90\% & 78\% & 98\% & 98\% & 88\% & 88\% & 92\% & 80\% & 62\% & 92\% & 94\% \\  
            
            \midrule[\heavyrulewidth] 
            
            \multirow{2}{*}{\textbf{Task Split}} & \multicolumn{10}{c|}{\textbf{LIBERO-Goal}} & \multicolumn{10}{c}{\textbf{LIBERO-Long}} \\
            \cmidrule(lr){2-11} \cmidrule(lr){12-21}
             & \textbf{T-1} & \textbf{T-2} & \textbf{T-3} & \textbf{T-4} & \textbf{T-5} & \textbf{T-6} & \textbf{T-7} & \textbf{T-8} & \textbf{T-9} & \textbf{T-10} & \textbf{T-1} & \textbf{T-2} & \textbf{T-3} & \textbf{T-4} & \textbf{T-5} & \textbf{T-6} & \textbf{T-7} & \textbf{T-8} & \textbf{T-9} & \textbf{T-10} \\
            \midrule
            
            \multicolumn{21}{l}{\cellcolor{gray!20}\textit{\textbf{Base Task Stage}}} \\
            \midrule
            Base Task 1-6 & 100\% & 98\% & 94\% & 86\% & 94\% & 96\% & -- & -- & -- & -- & 78\% & 86\% & 92\% & 96\% & 88\% & 92\% & -- & -- & -- & -- \\
            \midrule
            
            \multicolumn{21}{l}{\cellcolor{gray!20}\textit{\textbf{LifeLong Learning Stage}}} \\
            \midrule
            New Task 7  & 90\% & 90\% & 86\% & 88\% & 98\% & 94\% & 72\% & -- & -- & -- & 58\% & 78\% & 74\% & 94\% & 44\% & 86\% & 36\% & -- & -- & -- \\
            New Task 8  & 88\% & 96\% & 90\% & 76\% & 96\% & 90\% & 80\% & 100\% & -- & -- & 52\% & 70\% & 60\% & 84\% & 30\% & 80\% & 44\% & 82\% & -- & -- \\
            New Task 9  & 94\% & 94\% & 98\% & 80\% & 94\% & 96\% & 82\% & 98\% & 100\% & -- & 60\% & 70\% & 82\% & 88\% & 44\% & 94\% & 50\% & 80\% & 34\% & -- \\
            \rowcolor[HTML]{ECF4FF} 
            \textbf{New Task 10} & 86\% & 100\% & 92\% & 80\% & 98\% & 90\% & 78\% & 96\% & 86\% & 84\% & 58\% & 80\% & 70\% & 82\% & 38\% & 88\% & 38\% & 76\% & 18\% & 58\% \\  
            
            \bottomrule\bottomrule
        \end{tabular}%
    }
    \vspace{-0.2cm}
\end{table*}
\begin{table}[tp]
    \centering
    \caption{Detailed continual learning results in real-world experiments.}
    \label{tab:detailed_continual_learning_real_world}
    \vspace{-0.1cm}
    \large
    \setlength{\tabcolsep}{3pt} 
    \renewcommand{\arraystretch}{1.}     
    \resizebox{\linewidth}{!}{%
        \begin{tabular}{l|cccc}
            \toprule\toprule
            \textbf{Task Split} & \textbf{Pick Banana} & \textbf{Pick Bread} & \textbf{Pull Drawer} & \textbf{Hang Chinese Knot} \\
            \midrule
            \multicolumn{5}{l}{\cellcolor{gray!20}\textit{\textbf{LifeLong Learning Stage}}} \\
            \midrule
            New Task 1 & 85\% & -- & -- & -- \\
            New Task 2 & 80\% & 75\% & -- & -- \\
            New Task 3 & 70\% & 65\% & 100\% & -- \\
            \rowcolor[HTML]{ECF4FF} 
            \textbf{New Task 4} & 70\% & 70\% & 95\% & 60\% \\  
            \bottomrule\bottomrule
        \end{tabular}%
    }
    \vspace{-0.3cm}
\end{table}
Additionally, Table~\ref{tab:detailed_continual_learning_real_world} details the evaluation results of the real-world experiments.
In particular, our method achieves a 100\% success rate on the Pull Drawer task with only 20 demonstrations, demonstrating its superior plasticity and stability.
Nevertheless, for the deformable task (Hang Chinese Knot), the success rate remains at 60\%, suggesting the need for further improvement.

\subsubsection{Further Analysis of Continual Learning}
To further validate the effectiveness of LifeLong-RFT in learning extended task sequences, we conduct lifelong learning experiments across 10 tasks on the LIBERO-Goal suite.
Specifically, the training for each new task utilizes only 10 demonstrations, with 5 demonstrations per previous task preserved for experience replay.
As demonstrated in Table~\ref{tab:further_continual_learning_libero_goal}, despite the dual challenges of an increasing number of new tasks and limited training samples, our method exhibits strong adaptability to new tasks (\eg, achieving a 100\% success rate on Task 8) while maintaining stability on prior knowledge.
\begin{table}[ht]
    \centering
    \caption{Continual learning performance on LIBERO-Goal during the lifelong learning stage.}
    \label{tab:further_continual_learning_libero_goal}
    \vspace{-0.1cm}
    \large
    \setlength{\tabcolsep}{3pt} 
    \renewcommand{\arraystretch}{1.}     
    \resizebox{\linewidth}{!}{%
        \begin{tabular}{l|cccccccccc}
            \toprule\toprule
            \textbf{Task Split} & \textbf{T-1} & \textbf{T-2} & \textbf{T-3} & \textbf{T-4} & \textbf{T-5} & \textbf{T-6} & \textbf{T-7} & \textbf{T-8} & \textbf{T-9} & \textbf{T-10} \\
            \midrule
            \multicolumn{11}{l}{\cellcolor{gray!20}\textit{\textbf{LifeLong Learning Stage}}} \\
            \midrule
            New Task 1 & 48\% & -- & -- & -- & -- & -- & -- & -- & -- & -- \\
            New Task 2 & 44\% & 76\% & -- & -- & -- & -- & -- & -- & -- & -- \\
            New Task 3 & 30\% & 48\% & 94\% & -- & -- & -- & -- & -- & -- & -- \\
            New Task 4 & 54\% & 56\% & 96\% & 86\% & -- & -- & -- & -- & -- & -- \\
            New Task 5 & 48\% & 56\% & 98\% & 82\% & 98\% & -- & -- & -- & -- & -- \\

            New Task 6 & 38\% & 74\% & 88\% & 76\% & 72\% & 90\% & -- & -- & -- & -- \\
            New Task 7  & 40\% & 72\% & 54\% & 78\% & 76\% & 76\% & 54\% & -- & -- & -- \\
            New Task 8  & 44\% & 76\% & 68\% & 62\% & 80\% & 72\% & 60\% & 100\% & -- & -- \\
            New Task 9  & 26\% & 84\% & 88\% & 74\% & 96\% & 86\% & 60\% & 100\% & 96\% & -- \\
                        
            \rowcolor[HTML]{ECF4FF} 
            \textbf{New Task 10} & 34\% & 76\% & 88\% & 70\% & 94\% & 80\% & 64\% & 100\% & 98\% & 70\% \\  
            \bottomrule\bottomrule
        \end{tabular}%
    }
    \vspace{-0.3cm}
\end{table}

\subsection{Analysis of Reward Combinations within the Multi-Dimensional Process Reward}
\begin{figure}[ht]
    \centering
    \includegraphics[width=\linewidth]{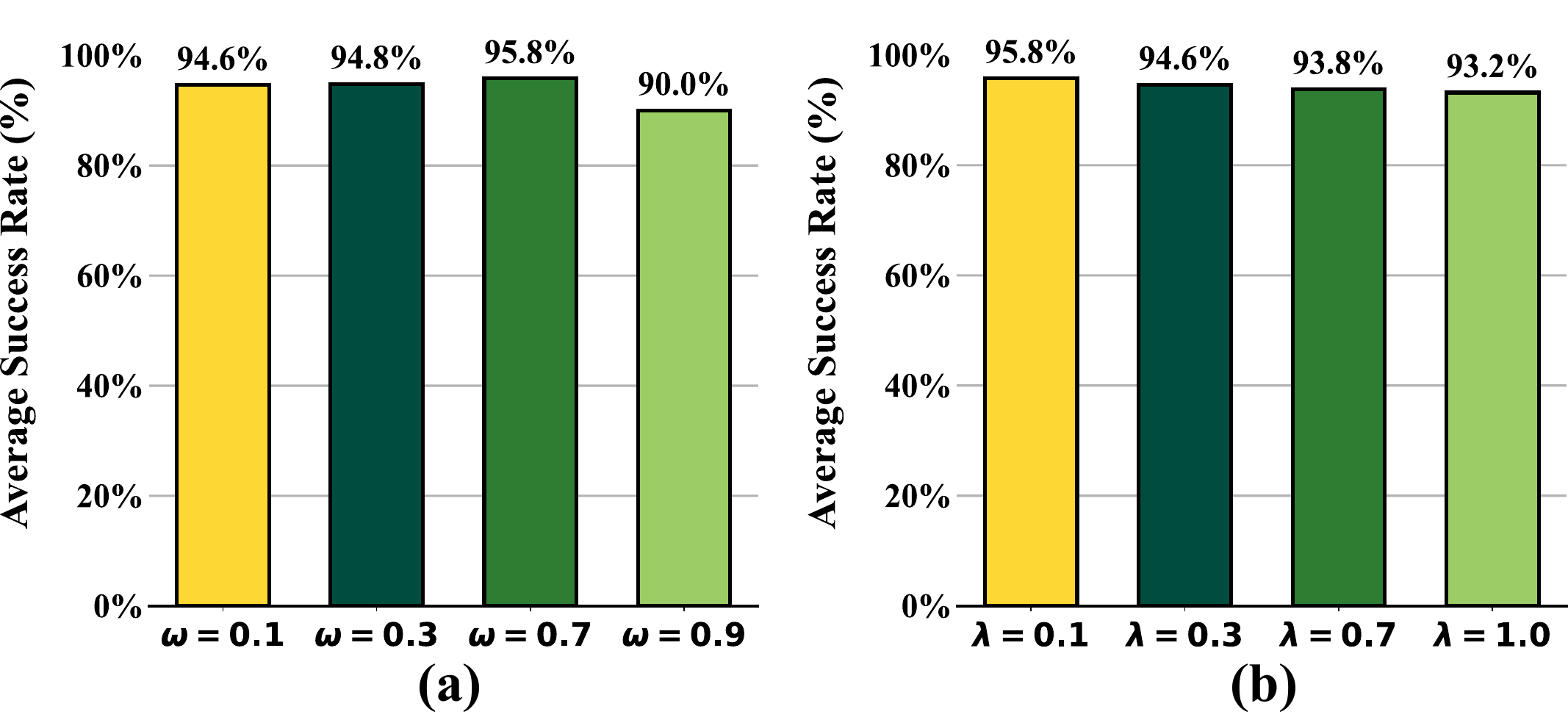}
    \captionof{figure}{Ablation study on the reward combination weights.
    }
\label{fig:Ablation_Hyper}
\vspace{-0.2cm}
\end{figure}

To evaluate the impact of reward combination weights within the multi-dimensional process reward, we conduct multi-task learning experiments on LIBERO-Goal, performing a detailed parameter sensitivity analysis of $\omega$ and $\lambda$.
As shown in Fig.~\ref{fig:Ablation_Hyper} (a), the model maintains comparable performance with $\omega$ values of 0.1, 0.3, and 0.7.
Specifically, when $\omega$ increases to 0.9, the weight of CTAR (\ie, $1-\omega = 0.1$) within the total reward significantly decreases, diminishing its guidance for model exploration and leading to a drop in the average success rate to 90.0\%.
Furthermore, the influence of the FCR-weighting hyperparameter $\lambda$ on model performance is illustrated in Fig.~\ref{fig:Ablation_Hyper} (b).
Experimental results demonstrate that our method also exhibits strong robustness to variations in this parameter.
In particular, we set $\lambda$ to 0.1, achieving optimal model performance.

\subsection{Visualization of Training Process}
\begin{figure}[ht]
    \centering
    \includegraphics[width=\linewidth]{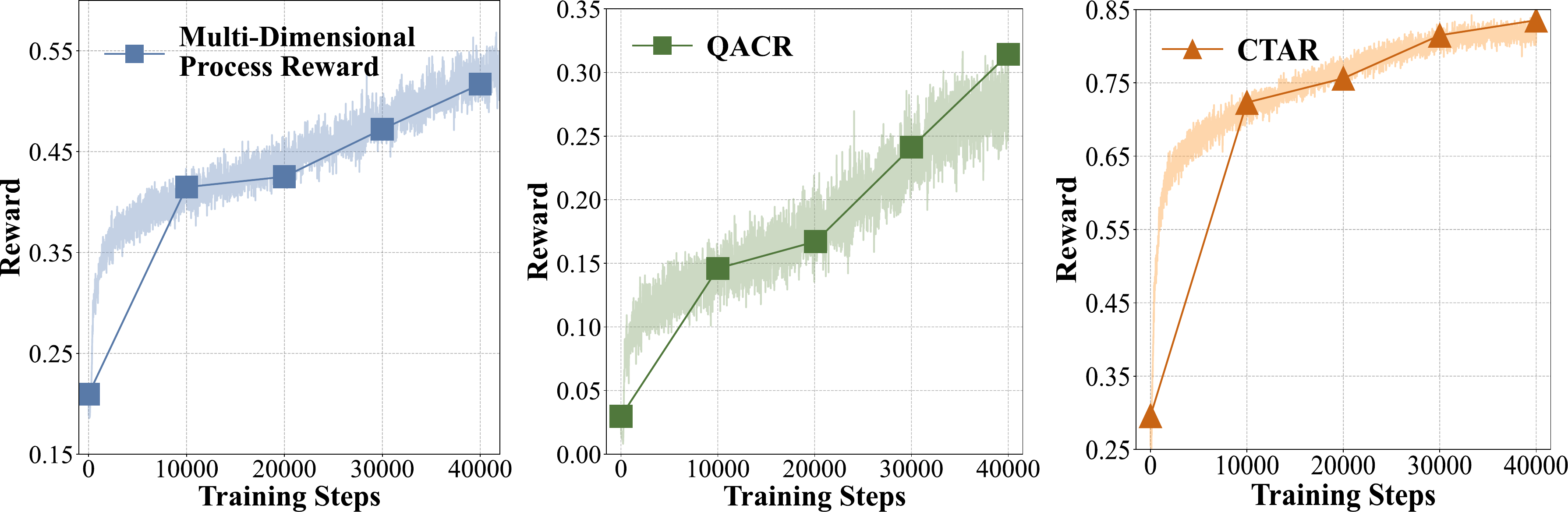}
    \captionof{figure}{Representative reward curves during the training phase. The visualizations illustrate the training evolution of (a) the Multi-Dimensional Process Reward, (b) QACR, and (c) CTAR.
    }
\label{fig:Reward_Curve}
\vspace{-0.3cm}
\end{figure}

To intuitively demonstrate the effectiveness of our proposed rewards during the reinforcement fine-tuning phase, Fig.~\ref{fig:Reward_Curve} presents the multi-task learning dynamics on LIBERO-Goal.
As shown in Fig.~\ref{fig:Reward_Curve} (a), the multi-dimensional process reward, comprising QACR, CTAR, and FCR, exhibits a continuous growth trend during training, confirming that it achieves synergistic optimization of the policy across multiple dimensions.
Furthermore, Fig.~\ref{fig:Reward_Curve} (b) and (c) illustrate that QACR and CTAR maintain consistent growth, indicating that they effectively incentivize the model to achieve precise manipulation.
\subsection{Generalization across Discrete Action Models}
To further validate the generalization of our approach across different discrete VLA models, we conduct comprehensive experiments built upon RoboBrain-X0~\cite{RoboBrain1.0}.
Table~\ref{tab:Robobrain_x0_simulation_continual_learning} demonstrates that LifeLong-RFT exhibits significant superiority across all four task suites.
Specifically, on LIBERO-Spatial, our approach yields a 12.2\% improvement in FWT and a 13.3\% reduction in NBT, demonstrating its capacity to efficiently acquire new skills while preserving previously learned knowledge.
\begin{table}[ht]
\centering
\caption{Continual learning performance of RoboBrain-X0 on LIBERO.}
\vspace{-0.1cm}
\label{tab:Robobrain_x0_simulation_continual_learning}
\scriptsize
\renewcommand{\arraystretch}{1.}
\renewcommand\theadgape{}
\newcommand{\pdm}[1]{{\color{black!60} #1}}
\definecolor{brickred}{HTML}{800000}

\resizebox{0.9\linewidth}{!}{%
\begin{tabular}{ll|cc >{\columncolor{blue!5}[2pt][\tabcolsep]}c}
\toprule\toprule
\multirow{2}{*}{\textbf{Task Split}} & \multirow{2}{*}{\textbf{Metrics}} & \multicolumn{3}{c}{\textbf{RoboBrain-X0}} \\
\cmidrule(lr){3-5}
& & \pdm{SFT} & \textbf{RFT (Ours)} & \multicolumn{1}{c}{\thead{$\boldsymbol{\Delta}$}} \\
\midrule

\multirow{3}{*}{LIBERO-Object}
 & FWT ($\uparrow$) & 86.6 & 94.0 & \textcolor{blue}{+7.4} \\
 & NBT ($\downarrow$) & 16.1 & 1.5 & \textcolor{brickred}{-14.6} \\
 \rowcolor[HTML]{ECF4FF}
 & \textbf{AUC} ($\uparrow$) & 74.7 & \textbf{92.4} & \cellcolor{blue!5}\textcolor{blue}{\textbf{+17.7}} \\
\midrule

\multirow{3}{*}{LIBERO-Spatial}
 & FWT ($\uparrow$) & 80.4 & 92.6 & \textcolor{blue}{+12.2} \\
 & NBT ($\downarrow$) & 21.0 & 7.7 & \textcolor{brickred}{-13.3} \\
 \rowcolor[HTML]{ECF4FF}
 & \textbf{AUC} ($\uparrow$) & 64.5 & \textbf{86.9} & \cellcolor{blue!5}\textcolor{blue}{\textbf{+22.4}} \\
\midrule

\multirow{3}{*}{LIBERO-Goal}
 & FWT ($\uparrow$) & 68.8 & 84.0 & \textcolor{blue}{+15.2} \\
 & NBT ($\downarrow$) & 16.0 & 13.1 & \textcolor{brickred}{-2.9} \\
 \rowcolor[HTML]{ECF4FF}
 & \textbf{AUC} ($\uparrow$) & 56.5 & \textbf{73.8} & \cellcolor{blue!5}\textcolor{blue}{\textbf{+17.3}} \\
\midrule

\multirow{3}{*}{LIBERO-Long}
 & FWT ($\uparrow$) & 62.8 & 64.4 & \textcolor{blue}{+1.6} \\
 & NBT ($\downarrow$) & 28.1 & 15.5 & \textcolor{brickred}{-12.6} \\
 \rowcolor[HTML]{ECF4FF}
 & \textbf{AUC} ($\uparrow$) & 40.8 & \textbf{52.0} & \cellcolor{blue!5}\textcolor{blue}{\textbf{+11.2}} \\
\bottomrule\bottomrule
\end{tabular}
}
\end{table}

\subsection{Real-World Case Studies}
To qualitatively analyze the performance of our method in real-world experiments, this section presents representative examples of execution across four tasks.

\subsubsection{Pick Banana}
\begin{figure*}[tp]
    \centering
    \includegraphics[width=0.9\linewidth]{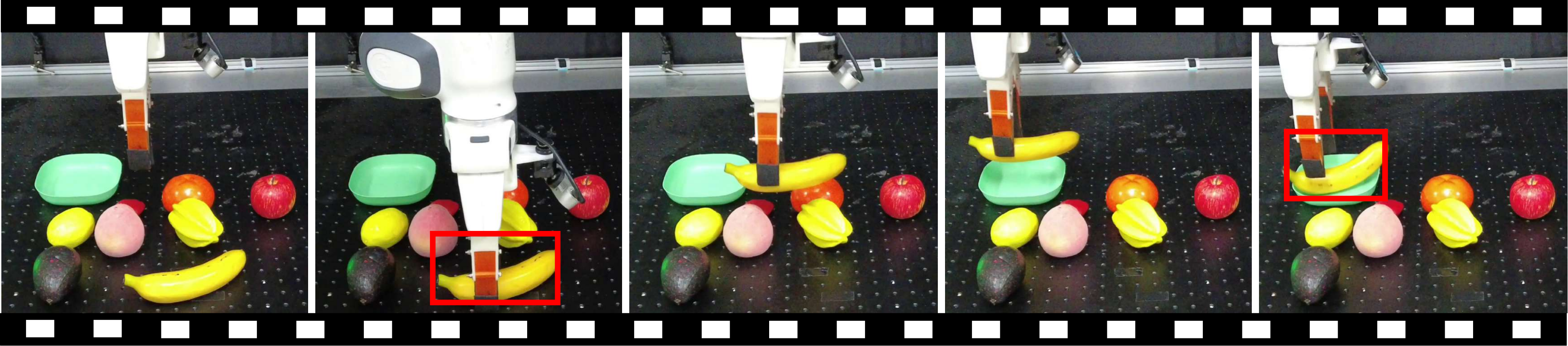}
    \captionof{figure}{A representative execution of the \textit{Pick Banana} task.
    }
\label{fig:case_banana}
\vspace{-0.4cm}
\end{figure*}

As illustrated in Fig.~\ref{fig:case_banana}, the task requires the model to accurately identify and grasp the banana from a cluttered scene containing various fruits, and subsequently place it stably into the blue plate.
Notably, our method effectively overcomes interference from distractor objects and robustly completes the pick-and-place task.

\subsubsection{Pick Bread}

\begin{figure*}[tp]
    \centering
    \includegraphics[width=0.9\linewidth]{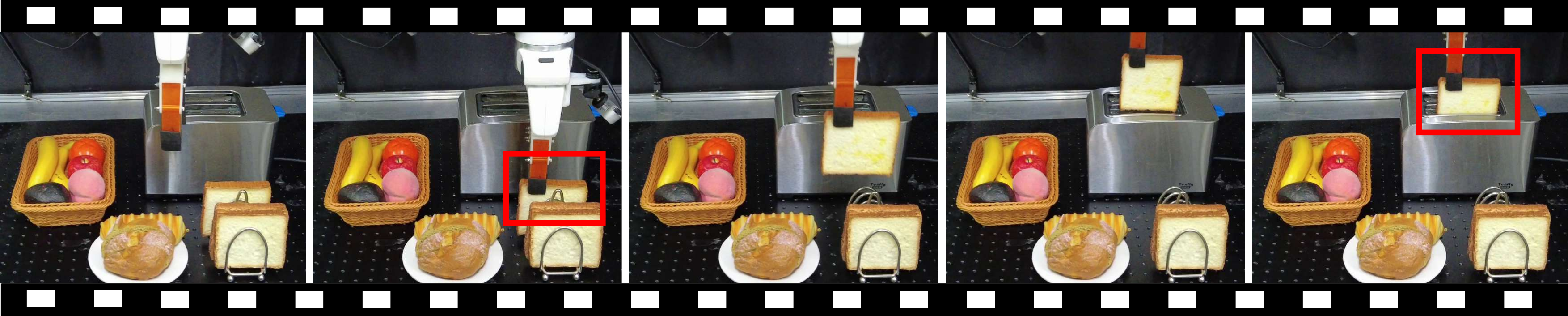}
    \captionof{figure}{A representative execution of the \textit{Pick Bread} task.
    }
\label{fig:case_bread}
\vspace{-0.4cm}
\end{figure*}

Fig.~\ref{fig:case_bread} demonstrates a representative execution of the Pick Bread task.
The core challenge of this task lies in the precise insertion of the bread into the narrow toaster slot.
The illustrated examples indicate that the model fine-tuned with LifeLong-RFT exhibits strong fine-grained manipulation capabilities, successfully completing this task.

\subsubsection{Pull Drawer}
\begin{figure*}[tp]
    \centering
    \includegraphics[width=0.9\linewidth]{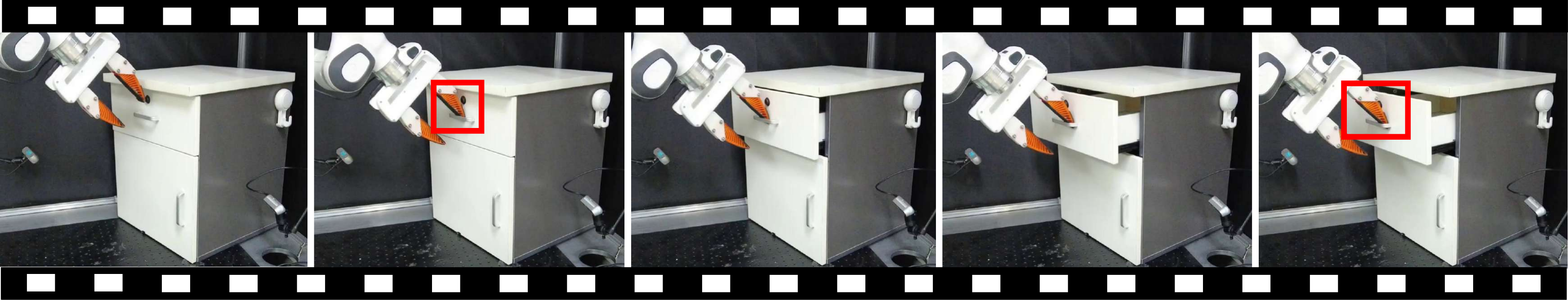}
    \captionof{figure}{
    A representative execution of the \textit{Pull Drawer} task.
    }
\label{fig:case_drawer}
\vspace{-0.4cm}
\end{figure*}

As shown in Fig.~\ref{fig:case_drawer}, the Pull Drawer task involves interacting with an articulated object, requiring the model to accurately grasp the handle and pull the drawer.
The primary difficulty stems from the requirement for strict coordination between the end-effector and the drawer's linear motion to avoid jamming.
Specifically, our approach demonstrates robust manipulation in constrained environments.
\subsubsection{Hang Chinese Knot}
\begin{figure*}[tp]
    \centering
    \includegraphics[width=0.9\linewidth]{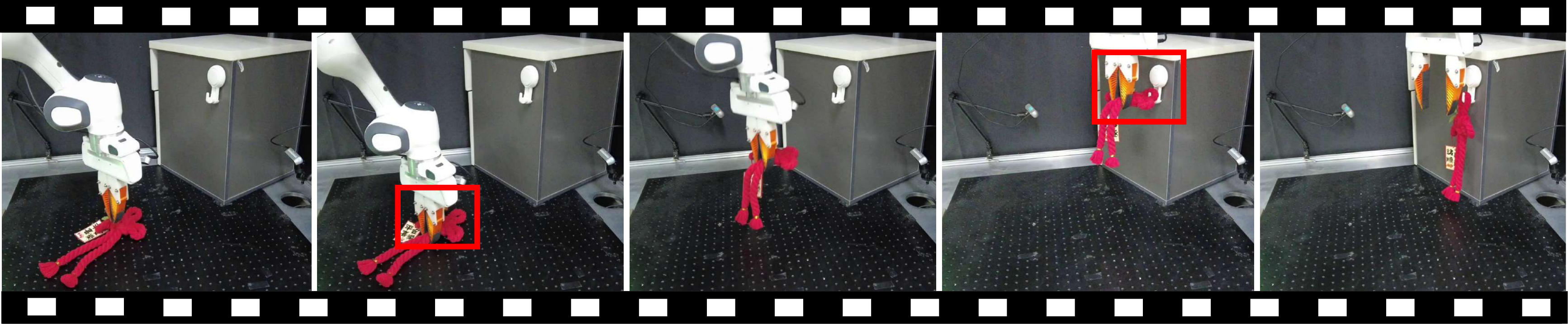}
    \captionof{figure}{A representative execution of the \textit{Hang Chinese Knot} task.
    }
\label{fig:case_knot}
\vspace{-0.4cm}
\end{figure*}

Fig.~\ref{fig:case_knot} illustrates the execution of the Hang Chinese Knot task, which centers on manipulating a deformable object.
The goal is to grasp the knot from the table and suspend it onto a cabinet-mounted hook.
This task necessitates superior fine-grained manipulation skills, enabling the model to execute the hanging operation while adapting to the dynamic deformations of the Chinese knot.
While our method demonstrates significant effectiveness, it also exhibits certain limitations, offering directions for future research.
%

%
%

\end{document}